\pdfoutput=1

\documentclass[11pt]{article}
\usepackage{amsmath}

\usepackage[final]{acl}

\usepackage{times}
\usepackage{latexsym}
\usepackage{amssymb}
\usepackage{multirow}

\usepackage[T1]{fontenc}

\usepackage[utf8]{inputenc}

\usepackage{microtype}

\usepackage{inconsolata}

\usepackage{graphicx}

\usepackage{subcaption}

\usepackage{colortbl}
\usepackage{booktabs}

\usepackage{fancyvrb}
\usepackage{tcolorbox}
\usepackage{verbatim}

\definecolor{promptbg}{rgb}{0.95,0.95,0.95}
\definecolor{promptframe}{rgb}{0.6,0.6,0.6}

\newtcolorbox{promptbox}[1][]{
    colback=promptbg,
    colframe=promptframe,
    boxrule=1pt,
    arc=3pt,
    boxsep=5pt,
    left=10pt,right=10pt,top=10pt,bottom=10pt,
    fonttitle=\bfseries,
    title={#1}
}

\title{Contrastive Decoding Mitigates Score Range Bias in LLM-as-a-Judge}

\author{Yoshinari Fujinuma \\
  Patronus AI \\
  \texttt{fujinumay@gmail.com}
  }

\begin{document}
\maketitle
\begin{abstract}
Large Language Models (LLMs) are commonly used as evaluators in various applications, but the reliability of the outcomes remains a challenge. One such challenge is using LLMs-as-judges for direct assessment, i.e., assigning scores from a specified range without any references. Focusing on summarization, we first show that this challenge stems from LLM judge outputs being associated with score range bias, i.e., LLM judge outputs are highly sensitive to pre-defined score ranges. We also show that similar biases exist among models from the same family. We then mitigate this bias through contrastive decoding, achieving up to $11.7\%$ relative improvement on average in Spearman correlation with human judgments across different score ranges.

\end{abstract}

\section{Introduction}
Large Language Model (LLM) judges have become an integral component of the evaluation ecosystem~\citep{lin-etal-2022-truthfulqa,chiang-lee-2023-large,bubeck2023sparksartificialgeneralintelligence}. In evaluations ranging from direct assessment, where judges evaluate individual outputs by assigning scores~\citep{liu-etal-2023-g}, to pairwise comparisons, where judges compare two outputs and determine which is superior~\citep{zheng2023judging,ye2024justiceprejudicequantifyingbiases}, using LLM as a judge is increasingly deployed to provide automatic, scalable, and cost-effective evaluation across diverse tasks. However, the reliability of such evaluations faces significant challenges, particularly when models assess their own outputs~\citep{zheng2023judging} or those from the same model family~\citep{goel2025greatmodelsthinkalike}. 
These biases constrain the set of models that can be reliably employed as LLM-as-a-judge for evaluation.
But could there be any other biases hidden when using LLMs as judges?

\begin{figure}[t]
    \centering
    \includegraphics[width=0.45\textwidth]{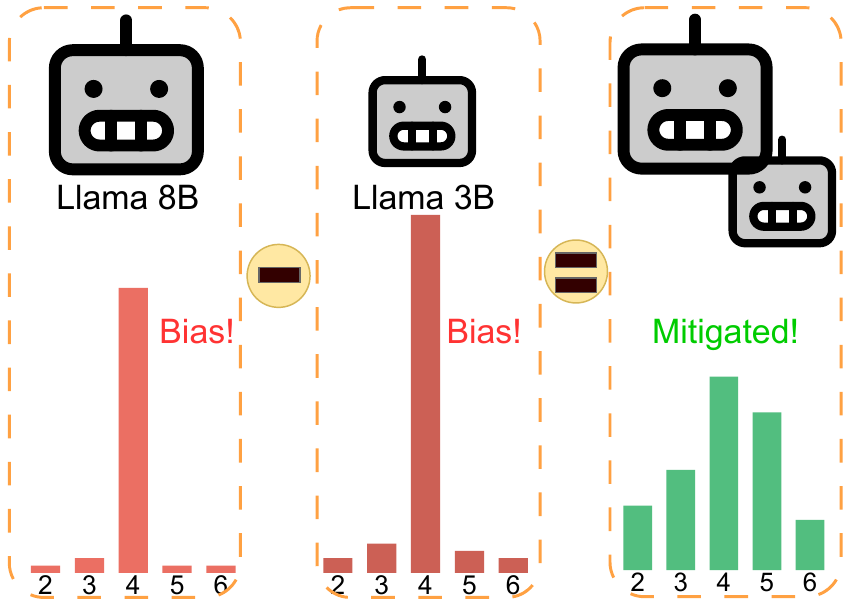}
    \caption{Overview of score range bias in 2-4 range and how contrastive decoding mitigates it through canceling out similar bias across models from the same family.}
    \label{fig:overview}
\end{figure}

We reveal another bias in LLM judge outputs, namely {\it score range biases}, where LLM judge outputs are sensitive to the shift in scores ranges, a phenomenon motivated by prior findings that LLMs struggle in simple arithmetic tasks~\citep{nogueira2021limitations,gambardella-etal-2024-language}.
Upon identifying such biases, we also explore a mitigation strategy by connecting recent work on contrastive decoding~\citep{li-etal-2023-contrastive,obrien2023contrastivedecodingimprovesreasoning} and family-enhancement bias~\citep{goel2025greatmodelsthinkalike}, aiming to cancel out similar score range biases encoded across the models from the same family.

Our summarized contributions are as follows:
\begin{itemize}
  \item We first show that LLM judges have {\it score range bias} -- a bias observed across different model sizes and families (Llama-3 and Qwen-2.5) when judging on direct assessment.
  \item We then show that contrastive decoding, motivated by the observation of similar score range biases shared across models from the same family, successfully mitigates these biases.
\end{itemize}

\section{Related Work}
We now review the related work on LLM judges focusing on the judge tasks and their biases.

\paragraph{LLM Judge Tasks} LLM judge tasks fall into two categories: {\it direct} assessment~\citep{jones-etal-2024-multi,li2024generative,zhu2025judgelm} and {\it pairwise} assessment~\citep{zheng2023judging,ye2024justiceprejudicequantifyingbiases}. Direct assessment~\citep{liu-etal-2023-g} involves assigning numerical ratings to a single output example. 
In pairwise assessment, LLM judges show higher correlation with human preferences than direct assessment~\citep{liu2024aligning}, supporting that the challenge remain in direct assessment, and we therefore focus on experimenting on direct assessment.

\paragraph{LLM Judge Biases}
One known bias in LLM judges is self-enhancement bias i.e., the tendency to favor their own output~\citep{liu-etal-2023-g,zheng2023judging,ye2024justiceprejudicequantifyingbiases} even in proprietary models like GPT-4~\citep{wataoka2024selfpreference}. 
Extending beyond self-enhancement bias, \citet{goel2025greatmodelsthinkalike} reported a family enhancement bias where models favor outputs from the same model family. 


%
%

\section{Analysis and Mitigation}
\subsection{Identifying Score Range Bias}

An ideal LLM judge should maintain consistent correlation with human judgments across shifted ranges, as they represent the same 5-point scale (e.g., 0-4, 1-5, 2-6, 3-7). Failure to do so indicates \textbf{score range bias}: models exhibit different correlations depending on the score range used, even when evaluating identical content. This bias outputs skewed distributions where models, for example, favor specific scores. 
We hypothesize that these biases can be mitigated with contrastive decoding, which cancels out shared biases.


\subsection{Mitigation by Contrastive Decoding}

Contrastive decoding~\citep{li-etal-2023-contrastive} modifies the model outputs by using two models: a main model and an assistant model.
Given the next token probability of a main model $p_{\text{main}}$ and an assistant model $p_{\text{asst}}$, 
the final adjusted score is calculated by subtracting the weighted $p_{\text{asst}}$ from $p_{\text{main}}$ i.e.,
\begin{equation}
\label{eq:constrastive}
\log p_{\text{main}} - \lambda \log p_{\text{asst}}
\end{equation}
where $\lambda \in \mathbb{R}$ is the hyperparameter to control the magnitude of assistant model and logit $e_i$ of token $i$ is controlled by temperature $t > 0$ i.e.,
$
p_{\text{Asst}} = \frac{e_i/t} {\sum_j e_j/t}.
$
In contrast to~\citet{li-etal-2023-contrastive}, we include $\lambda$ to align the logit distribution between the two models motivated by our analysis in \S\ref{sec:analysis}.

\section{Experiments}
We focus on direct assessment on summarization since prior work reported that LLM judges fall short~\citep{ye2024justiceprejudicequantifyingbiases} and they are commonly evaluated on summarization~\citep{panickssery2024llm}.

\subsection{Setup}
\paragraph{Task and Metrics} We focus on the summarization task where LLM judges are commonly used~\cite{liu-etal-2023-g,panickssery2024llm}.
The correlations between human annotations  are measured using three metrics: Pearson, Spearman, and Kendall correlations.

\paragraph{Score Scale and Ranges} We use the 5 points Likert scale~\citep{likert1932technique} on different score ranges (0-4, 1-5, 2-6, 3-7)\footnote{We stopped at 7 inspired by~\citet{likert1932technique} showing high correlation between 5 points (1-5) and 7 points (1-7) results.}.
If output score parsing fails, we set to the lowest score following~\citet{liu-etal-2023-g} and if the parsed score exceeds the maximum, we clamp to the highest score in the range.

\paragraph{Models}
We experiment on two model families.\footnote{We leave models like Prometheus~\citep{kim-etal-2024-prometheus} specifically fineutuned on judge tasks as future work since multiple model sizes are not available for contrastive decoding and those models are finetuned towards 1-5 score range.
} 
For Llama-3 family~\citep{grattafiori2024llama3herdmodels}, we use \texttt{Llama-3.1-8B-Instruct} as the main model, \texttt{Llama-3.2-3B-Instruct} and \texttt{Llama-3.2-1B-Instruct} as the assistant model, and for Qwen2.5 family~\citep{qwen2025qwen25technicalreport}, we use   \texttt{Qwen-2.5-14B-Instruct} and \texttt{Qwen-2.5-7B-Instruct} as the main models, and \texttt{Qwen-2.5-3B-Instruct} as the assistant model. See Appendix~\ref{sec:appendix_prompt} for the prompt used.

\begin{figure}[t]
    \centering
    \begin{subfigure}{0.49\textwidth}
        \centering
        \includegraphics[width=\textwidth]{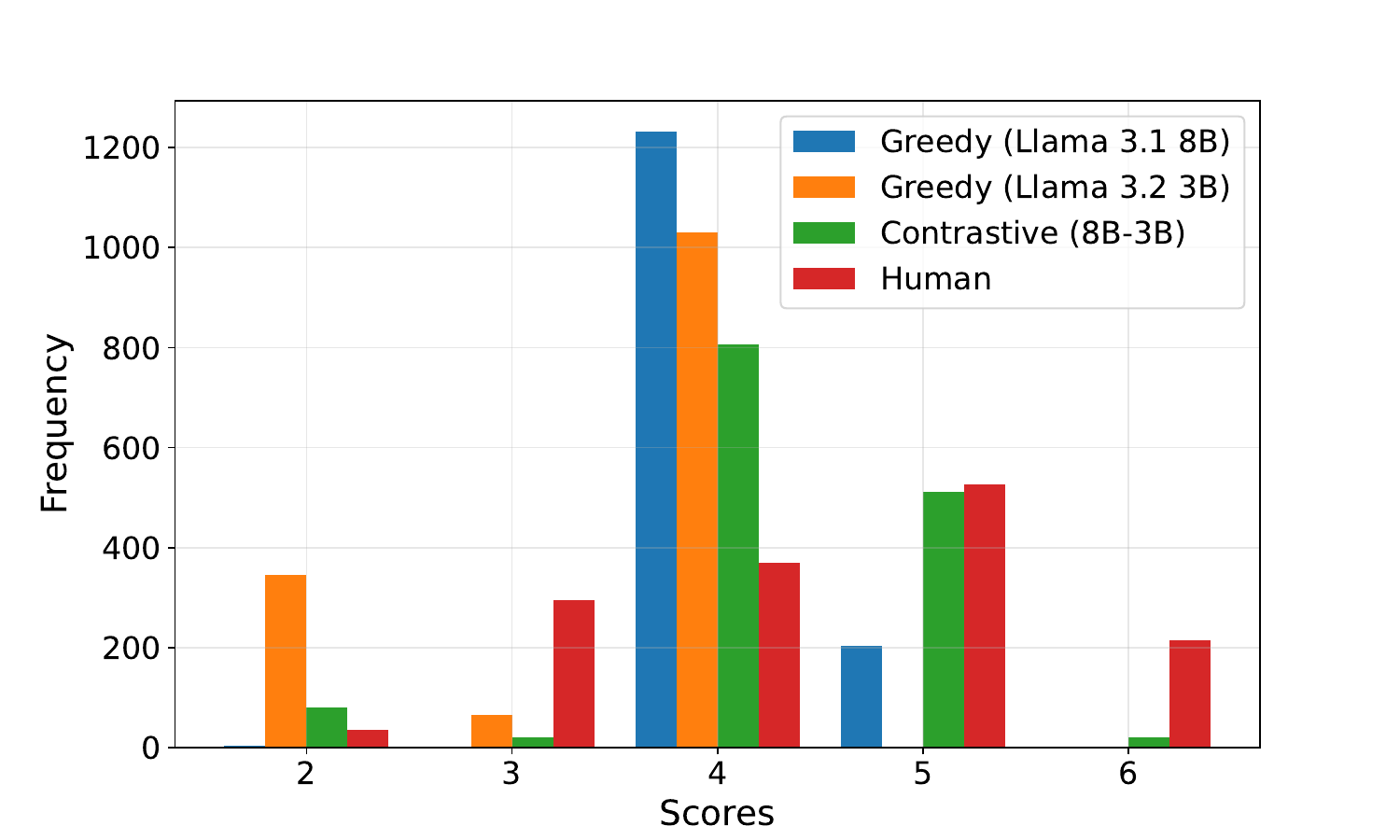}
        \caption{Llama-3 Family Results}
        \label{fig:score_dist_llama}
    \end{subfigure}
    \hfill
    \begin{subfigure}{0.49\textwidth}
        \centering
        \includegraphics[width=\textwidth]{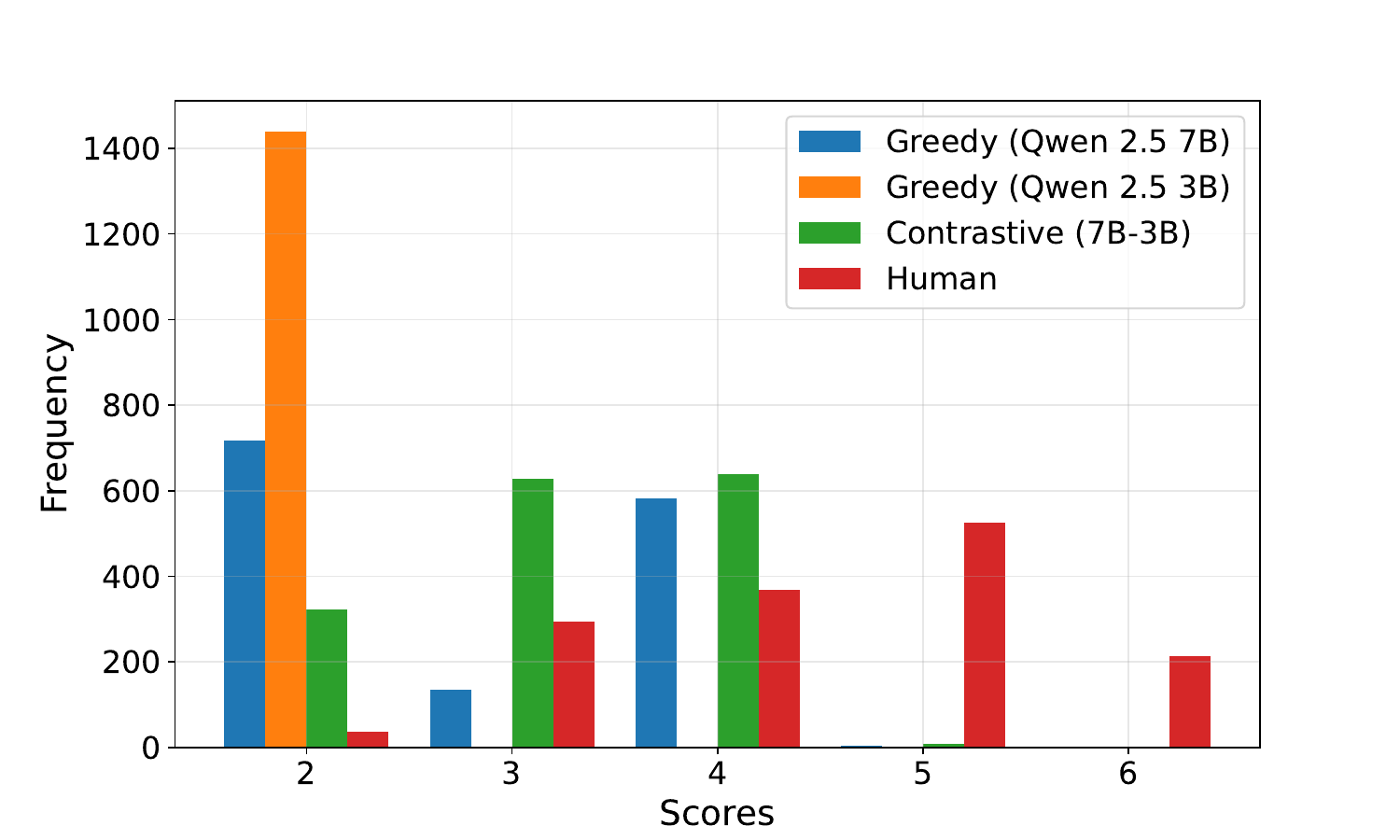}
        \caption{Qwen2.5 Family Results}
        \label{fig:score_dist_qwen}
    \end{subfigure}
    \caption{Coherence score distribution in 2-6 score range with greedy decoding, contrastive decoding, and human annotations. The greedy decoding outputs from both Llama 8B and 3B models are highly skewed towards outputting score of 4. Qwen2.5 3B (see also Figure~\ref{fig:logit_dist_qwen3b}) and 7B models are outputting score of 2 showing similar biases are encoded in these models.}
    \label{fig:score_distribution}
\end{figure}

\paragraph{Dataset}
We use SummEval~\citep{fabbri-etal-2021-summeval}, a summarization benchmark also used by \citet{liu-etal-2023-g} which contains 100 news articles where each article is associated with 16 summaries with human annotation scores, which sums up to 1600 summaries. $10\%$ of the news articles are used as the held out development set to conduct grid search.

We now first reveal the score range bias of LLM judges by evaluating correlation to human annotations in the same 5 points scale but with different score ranges, and then experiment on mitigating it.

\begin{figure}[t]
    \centering
    \begin{subfigure}{0.45\textwidth}
        \centering
        \includegraphics[width=\textwidth]{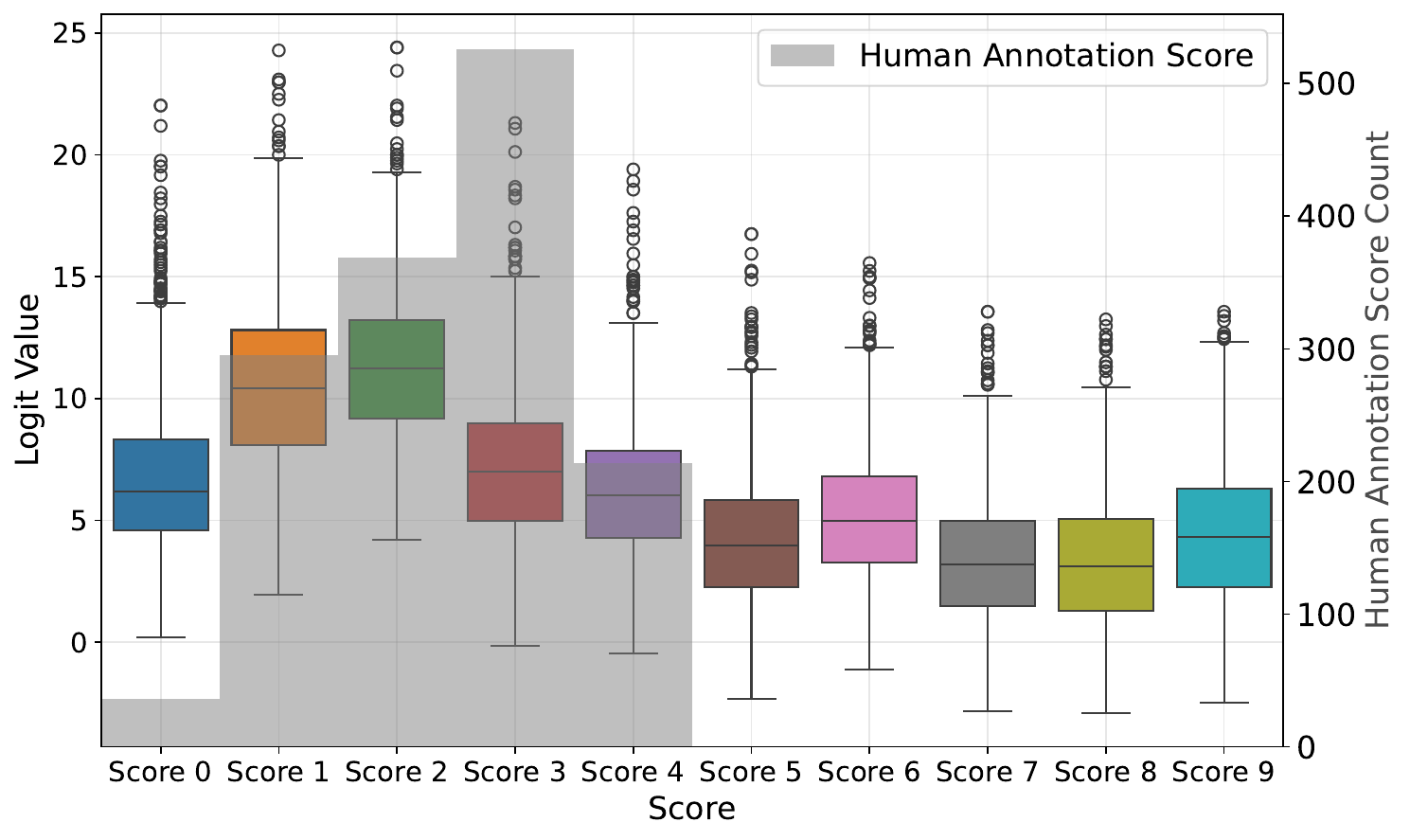}
        \caption{Qwen2.5 3B logit distribution}
        \label{fig:logit_dist_qwen3b}
    \end{subfigure}
    \hfill
    \begin{subfigure}{0.45\textwidth}
        \centering
        \includegraphics[width=\textwidth]{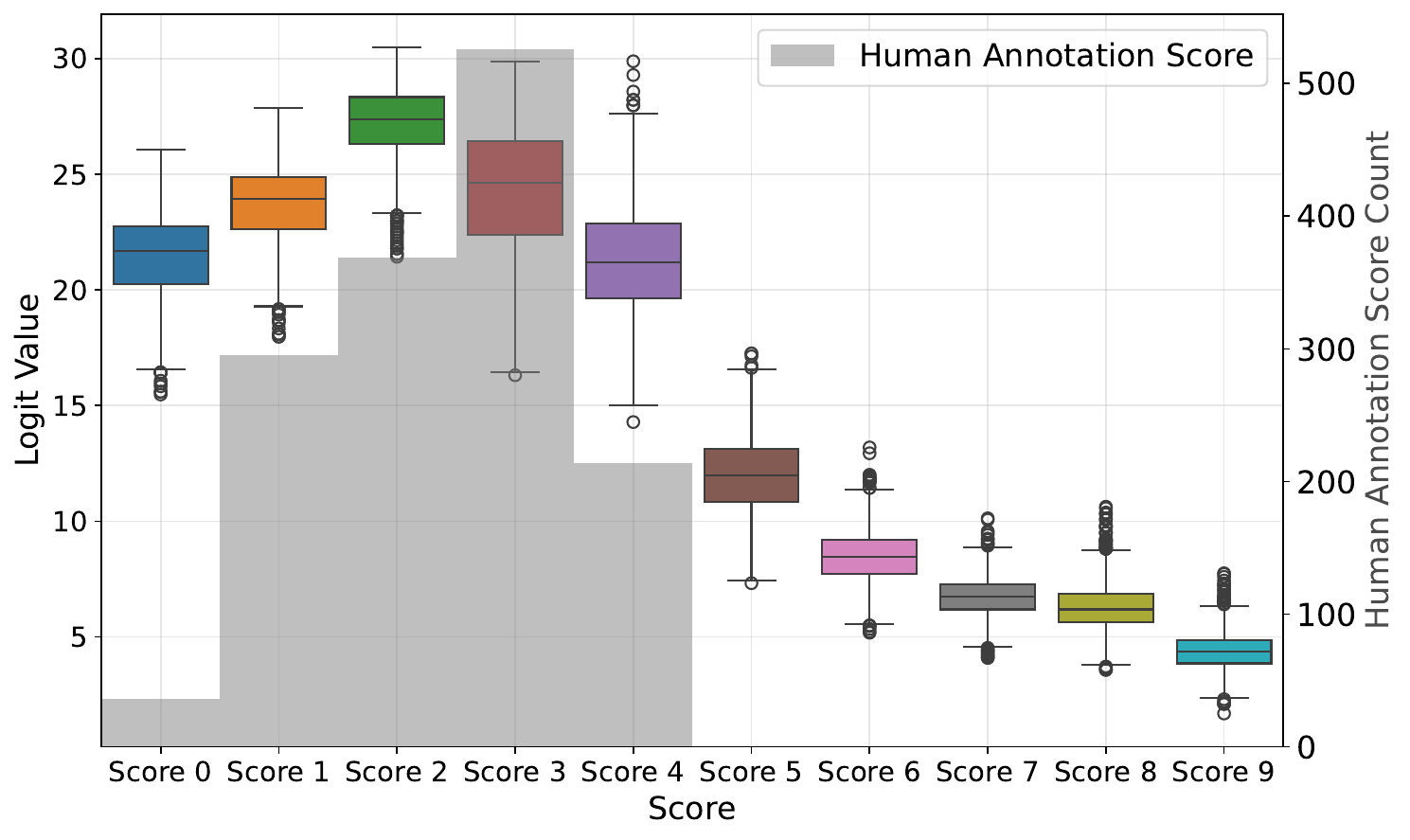}
        \caption{Qwen2.5 7B logit distribution}
        \label{fig:logit_dist_qwen7b}
    \end{subfigure}
    \hfill
    \begin{subfigure}{0.45\textwidth}
        \centering
        \includegraphics[width=\textwidth]{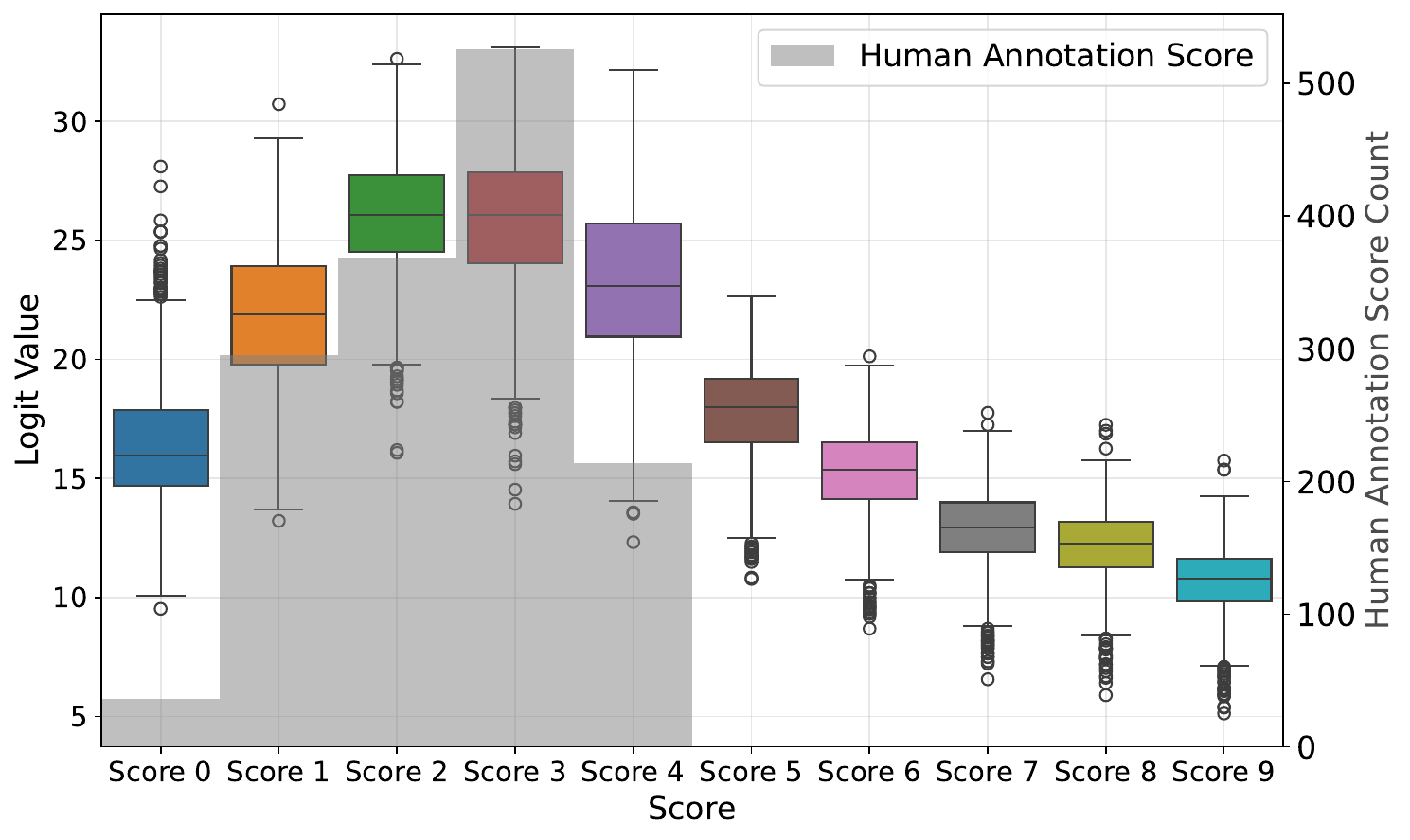}
        \caption{Qwen2.5 14B logit distribution}
        \label{fig:logit_dist_qwen14b}
    \end{subfigure}
    \caption{Logit distribution of the first output token in 0-4 score range. For 3B and 7B models, the logit of Score 2 is the highest while the logit of Score 3 gets higher and becomes closer to human judgements as the model size increase from 3B through 14B.}
    \label{fig:logit_distribution}
\end{figure}

\begin{table}[t]
\small
  \centering
  \label{tab:results_llama}
  \setlength{\tabcolsep}{4pt}
  \begin{tabular}{lllllllll}
    \toprule
    Model             & Range &  Pear. & Spear. & Kend.   \\
    \midrule                                                                             
    Llama 3.2-1B   & 0 to 4 & $.073_{.036}$ &  $.055_{.035}$ &  $.047_{.016}$ \\
    Llama 3.2-3B   & 0 to 4 & $.056_{.033}$ &  $.089_{.028}$  &  $.073_{.026}$ \\
    Llama 3.1-8B   & 0 to 4 & \underline{$.407_{.029}$} &  \underline{$.384_{.036}$}  &  \underline{$.327_{.022}$} \\
    Contra. (8B-1B) & 0 to 4 & $.384_{.037}$ & $.363_{.038}$ & $.309_{.025}$ \\
    Contra. (8B-3B) & 0 to 4 & $.380_{.030}$ & $.357_{.048}$ &  $.303_{.040}$ \\
    \midrule                                                                             
    Llama 3.2-1B & 1 to 5 &  $.000_{.000}$ &  $.000_{.000}$  &  $.000_{.000}$ \\
    Llama 3.2-3B  & 1 to 5 & $.151_{.036}$ &  $.201_{.042}$  &  $.163_{.016}$ \\
    Llama 3.1-8B  & 1 to 5 & $.338_{.036}$ &  $.309_{.042}$ &  $.262_{.029}$ \\
    Contra. (8B-1B)& 1 to 5 & \underline{$.357_{.021}$} & \underline{$.390_{.049}$} & \underline{$.327_{.030}$} \\
    Contra. (8B-3B)& 1 to 5 &  $.323_{.043}$ &  $.315_{.029}$  &  $.265_{.029}$ \\
    \midrule                                                                             
 
    Llama 3.2-1B  & 2 to 6 & $.000_{.000}$ &  $.000_{.000}$ & $.000_{.000}$ \\
    Llama 3.2-3B  & 2 to 6 & $.008_{.032}$ &  $.031_{.029}$ & $.026_{.027}$ \\
    Llama 3.1-8B  & 2 to 6 & $.262_{.029}$ &  $.257_{.025}$ &  $.220_{.015}$ \\
    Contra. (8B-1B)& 2 to 6 & \underline{$.344_{.038}$} & \underline{$.337_{.035}$} & \underline{$.288_{.034}$} \\
    Contra. (8B-3B)& 2 to 6 & $.302_{.049}$ &  $.305_{.034}$ &  $.254_{.023}$ \\
    \midrule                                                                             
    Llama 3.2-1B  & 3 to 7          &  -$.039_{.035}$ & -$.051_{.020}$ & -$.043_{.028}$ \\
    Llama 3.2-3B  & 3 to 7          & -$.007_{.035}$ &  -$.009_{.044}$ &  -$.008_{.027}$  \\
    Llama 3.1-8B  & 3 to 7          & \underline{\it $.445_{.027}$} &  \underline{\it $.426_{.030}$} &  \underline{\it $.352_{.030}$} \\
    Contra. (8B-1B)& 3 to 7          & $.425_{.030}$ &  $.408_{.033}$ &  $.341_{.019}$ \\
    Contra. (8B-3B)& 3 to 7          & $.442_{.018}$ &  $.419_{.035}$ &  $.347_{.023}$  \\
    \hline
    \multicolumn{5}{c}{\it Average across all score ranges}  \\
    Llama 3.2-1B  & & $.009$ & $.001$ & $.001$ \\
    Llama 3.2-3B  & & $.052$ & $.078$ & $.064$ \\
    Llama 3.1-8B  & & $.363$ & $.344$ & $.290$  \\
    Contra. (8B-1B)& & $\mathbf{.378}$ & $\mathbf{.375}$ & $\mathbf{.316}$ \\
    Contra. (8B-3B)& & $.362$ & $.349$ & $.292$ \\
    \bottomrule
  \end{tabular}
  \caption{\label{tab:contrastive_judge_corr_llama} Llama-3 family correlation to humans on summary coherence with 95\% confidence interval from bootstrap testing. Max correlation within score range are \underline{underlined} and max averages are {\bf bolded}. Maximal improvement is observed in the 2-6 score range.}
\end{table}

\begin{table}[ht]
\small
  \centering
  \label{tab:results_qwen}
  \setlength{\tabcolsep}{4pt}
  \begin{tabular}{lllllllll}
    \toprule
    Model             & Range &  Pear. & Spear. & Kend. \\
    \midrule
    Qwen2.5-3B      & 0 to 4 & -$.042_{.038}$ & -$.059_{.023}$ & -$.051_{.028}$ \\
    Qwen2.5-7B      & 0 to 4 & $.248_{.048}$ &  $.245_{.028}$ & $.209_{.018}$ \\
    Qwen2.5-14B     & 0 to 4 & $.435_{.041}$ &  $.452_{.019}$ & $.382_{.021}$ \\
    Contra. (7B-3B)  & 0 to 4 &  $.334_{.033}$ &  $.332_{.048}$ &  $.282_{.023}$ \\
    Contra. (14B-3B) & 0 to 4 & \underline{$.442_{.037}$} & \underline{$.458_{.031}$} & \underline{$.385_{.025}$} \\
    \midrule
    Qwen2.5-3B      & 1 to 5 & -$.044_{.042}$ & -$.056_{.040}$ & -$.048_{.015}$ \\
    Qwen2.5-7B      & 1 to 5 & $.382_{.043}$ & $.370_{.042}$ &  $.309_{.024}$ \\
    Qwen2.5-14B     & 1 to 5 & \underline{\it $.468_{.040}$} & $.468_{.026}$ & $.385_{.025}$ \\
    Contra. (7B-3B)  & 1 to 5 &  $.365_{.042}$ &  $.356_{.056}$ &  $.297_{.019}$ \\
    Contra. (14B-3B) & 1 to 5 &  $.463_{.039}$ &  \underline{\it $.470_{.041}$} & \underline{\it $.387_{.027}$} \\
    \midrule
    Qwen2.5-3B      & 2 to 6 & -$.010_{.000}$ & -$.013_{.000}$ & -$.011_{.000}$ \\
    Qwen2.5-7B      & 2 to 6 &  $.362_{.024}$ & $.355_{.036}$ &  $.293_{.025}$ \\
    Qwen2.5-14B     & 2 to 6 &  $.310_{.037}$ &  $.313_{.056}$ & $.260_{.026}$ \\
    Contra. (7B-3B)  & 2 to 6 & $.373_{.041}$ & $.364_{.034}$ & $.298_{.021}$  \\
    Contra. (14B-3B) & 2 to 6 & \underline{$.410_{.028}$} & \underline{$.426_{.035}$} &  \underline{$.359_{.040}$}  \\
    \midrule
    Qwen2.5-3B      & 3 to 7 & $.033_{.016}$ &  $.034_{.000}$  &  $.029_{.012}$ \\
    Qwen2.5-7B      & 3 to 7 & $.341_{.045}$ &  $.349_{.024}$ &  $.289_{.037}$ \\
    Qwen2.5-14B     & 3 to 7 & $.349_{.039}$ &  $.336_{.036}$ &  $.275_{.036}$ \\
    Contra. (7B-3B)  & 3 to 7 & $.351_{.019}$ &  $.357_{.034}$ &  $.298_{.027}$ \\
    Contra. (14B-3B) & 3 to 7 & \underline{$.413_{.027}$} & \underline{$.398_{.020}$} & \underline{$.327_{.029}$}  \\
    \hline
    \multicolumn{5}{c}{\it Average across all score ranges}  \\
    Qwen2.5-3B      & & -$.016$ & -$.024$ & -$.020$\\
    Qwen2.5-7B      & & $.333$ & $.330$ & $.275$\\
    Qwen2.5-14B     & & $.391$ & $.392$ & $.326$ \\
    Contra. (7B-3B)  & & $.356$ & $.352$ & $.294$ \\
    Contra. (14B-3B) & & $\mathbf{.432}$ & $\mathbf{.438}$ & $\mathbf{.365}$ \\
    \bottomrule
  \end{tabular}
  \caption{\label{tab:contrastive_judge_corr_qwen} Qwen-2.5 family correlation to humans on summary coherence with  95\% confidence interval from bootstrap testing.
  Max correlation within score range are \underline{underlined} and max averages are {\bf bolded}. Maximal improvement is observed in the 2-6 score range.}
\end{table}

\subsection{Reveal and Mitigate Score Range Biases}
\label{sec:analysis}

\paragraph{Similar score range biases exist across models from the same family}
We first analyze the distribution of the output scores in the 2 to 6 score range (Figure~\ref{fig:score_distribution}). Llama family models (3B and 8B) tend to output score of 4 (Figure~\ref{fig:score_dist_llama}) and Qwen 2.5 family models tend to output score of 2 (Figure~\ref{fig:score_dist_qwen}). 
By using contrastive decoding, these biases toward specific ranges are mitigated and making the score outputs closer to human annotations.

Upon analyzing the first output token logit distribution~(Figure~\ref{fig:logit_distribution}) of the Qwen family models in the 0-4 score range, Qwen-2.5 3B, 7B, 8B, and 14B models encode similar biases where Score 2 is the highest while the most frequent human annotation is Score 3. The bias towards Score 2 gradually decreases as the model size scales from 3B to 14B, but still remains even in the 14B model. Furthermore, the logit range in each model differs e.g., max logit in 3B $\approx 25$ (Figure~\ref{fig:logit_dist_qwen3b}), 7B $\approx 30$ (Figure~\ref{fig:logit_dist_qwen7b}), and 14B $\approx 34$, motivating the inclusion of $\lambda$ in Eq.~\ref{eq:constrastive} to align the logit distributions between these models. This bias on Score 2 in the 3B model helps decrease similar bias encoded in the 7B and 14B models when used as the assistant model.

As a result of score range bias, using Llama 3B or 7B with greedy decoding causes the lowest correlation in 2-6 score range (Table~\ref{tab:contrastive_judge_corr_llama}).
This trend is not limited to the Llama-3 family models and  it is also observed in the Qwen-2.5 family models (Table~\ref{tab:contrastive_judge_corr_qwen}). 
Focusing on greedy decoding, Qwen-2.5 family and Llama3.1-3B show clearer trend that the 1-5 score range shows the highest correlation among the experimented score ranges (7B, 1-5: .370, 14B, 1-5: .468), while Llama3.1-8B being the exception that 3 to 7 score range showing highest correlation among all score ranges. 
These outcomes further raises concern on applying LLM judges beyond the standard 1-5 range.

\paragraph{Contrastive Decoding is a Robust Mitigation Strategy across Different Score Ranges} 
 Table~\ref{tab:contrastive_judge_corr_llama} and \ref{tab:contrastive_judge_corr_qwen} further show  that contrastive decoding exhibits consistency in correlations across varying score ranges, addressing the score range bias observed. While using a single model suffers from decrease in correlation when score ranges are shifted, contrastive decoding maintains more stable correlations with human judgments regardless of the score ranges (Table~\ref{tab:contrastive_judge_corr_llama}). This robustness is evident in the 2-6 range, where contrastive decoding on Llama-3 family achieves a Pearson correlation of $.310$ (compared to $.168$ for Llama 3.2-3B and $.270$ for Llama 3.1-8B) and similar improvements in Spearman and Kendall correlations, also seen as $6.7\%$ relative improvement for Llama 8B ($.330 \rightarrow .352$) and $11.7\%$ for Qwen 14B ($.392 \rightarrow .438$) on average across all score ranges. 
 The stability across different scoring ranges enables search on optimal score ranges beyond the 1-5 range (e.g., 0-4 range showing the best correlation in summary relevance for Qwen family in Appendix~\ref{appendix:relevance_consistency}).

\paragraph{Does Assistant Model Choice Matter for Bias Mitigation?}

Table~\ref{tab:contrastive_judge_corr_llama} shows that the choice of assistant model slightly impacts correlations, with the 1B model marginally outperforming the 3B model. The 1B assistant achieves an average Spearman correlation of $.375$ compared to $.352$ for the 3B assistant.
Notably, using larger assistant models can degrade performance: our ablation study with Qwen 14B-7B (Appendix~\ref{appendix:assistant_size}) shows significant degradation in the 1-5 range (Spearman: .282 vs .470 with 3B assistant), showing that larger assistants penalize correct logits from the main model.

\section{Conclusion}
In this work, we analyze and experiment with LLM-as-a-judge on direct assessment, which reveals two key findings: First, LLM judges exhibit a score range bias across different model families and sizes with a tendency to favor specific scores regardless of the quality of the summaries. Second, we show that contrastive decoding effectively mitigates score range bias by leveraging the similar biases present in models from the same family. Our score range bias analysis framework can test arbitrary models and can help unlock the potential to expand beyond the standard 1-5 score range. 

\section*{Limitations}

\paragraph{Inference Time Compute}
Contrastive learning increases the test time compute due to running forward pass on two models rather than one. 
On the other hand, using a main model and an assistant model is very common in real world setup to speed up decoding with speculative decoding~\citep{speculative2023}, and therefore contrastive decoding can be used without an additional forward pass when speculative decoding is used.

\paragraph{Model Size}
Our experiments were limited to models with up to 14B parameters due to computational budget constraints. 

\paragraph{Language Coverage}
Our experiments are conducted only on English language, however, we have not exploited linguistic knowledge specific to English.

\paragraph{Task Coverage}
Our experiments are conducted on a summarization task as follows~\citet{liu-etal-2023-g} and \citet{panickssery2024llm}. However, we have conducted experiments on multiple dimensions of summarization metrics i.e., coherence, relevance (Appendix~\ref{appendix:relevance_consistency}), and consistency~(Appendix~\ref{appendix:relevance_consistency}) to confirm score range bias is not happening with one specific dimension.

\section*{Acknowledgments}
We sincerely thank the anonymous reviewers for their constructive feedback and insightful comments, which helped strengthen the analysis and improve the clarity of this work.

\bibliography{anthology,custom}

\appendix

\section{Judge Prompts}
\label{sec:appendix_prompt}
We use the following prompt experimented by~\citet{liu-etal-2023-g}.
\begin{promptbox}[Score Range \{min\_range\}-\{max\_range\} for Coherence]
\small
You will be given one summary written for a news article.
\\\\
Your task is to rate the summary on one metric.
\\\\
Please make sure you read and understand these instructions carefully. Please keep this document open while reviewing, and refer to it as needed.
\\\\
Evaluation Criteria:
\\\\
Coherence (\{min\_range\}-\{max\_range\}) - the collective quality of all sentences. We align this dimension with the DUC quality question of structure and coherence whereby "the summary should be well-structured and well-organized. The summary should not just be a heap of related information, but should build from sentence to a coherent body of information about a topic."
\\\\
Evaluation Steps:
\\\\
1. Read the news article carefully and identify the main topic and key points.\\
2. Read the summary and compare it to the news article. Check if the summary covers the main topic and key points of the news article, and if it presents them in a clear and logical order.\\
3. Assign a score for coherence on a scale of \{min\_range\} to \{max\_range\}, where \{min\_range\}  is the lowest and \{max\_range\} is the highest based on the Evaluation Criteria.
\\\\

Example:
\\\\

Source Text:
\\\\

\{\{Document]\}\}
\\\\

Summary:
\\\\

\{\{Summary\}\}
\\\\

Evaluation Form (scores ONLY):
\\\\

- Coherence:
\\\\

What is the coherence of the summary above? Provide only rating and no other text.
\end{promptbox}

\begin{promptbox}[Score Range \{min\_range\}-\{max\_range\} for Relevance]
\small
You will be given one summary written for a news article.
\\\\
Your task is to rate the summary on one metric.
\\\\
Please make sure you read and understand these instructions carefully. Please keep this document open while reviewing, and refer to it as needed.
\\\\
Evaluation Criteria:
\\\\
Relevance (\{min\_range\}-\{max\_range\}) - selection of important content from the source. The summary should include only important information from the source document. Annotators were instructed to penalize summaries which contained redundancies and excess information.
\\\\
Evaluation Steps:
\\\\
1. Read the summary and the source document carefully. \\
2. Compare the summary to the source document and identify the main points of the article. \\
3. Assess how well the summary covers the main points of the article, and how much irrelevant or redundant information it contains. \\
4. Assign a relevance score from \{min\_range\} to \{max\_range\}. \\
\\\\

Example:
\\\\

Source Text:
\\\\

\{\{Document]\}\}
\\\\

Summary:
\\\\

\{\{Summary\}\}
\\\\

Evaluation Form (scores ONLY):
\\\\

- Relevance:
\\\\

What is the relevance of the summary above? Provide only rating and no other text.
\end{promptbox}

\begin{promptbox}[Score Range \{min\_range\}-\{max\_range\} for Consistency]
\small
You will be given one summary written for a news article.
\\\\
Your task is to rate the summary on one metric.
\\\\
Please make sure you read and understand these instructions carefully. Please keep this document open while reviewing, and refer to it as needed.
\\\\
Evaluation Criteria:
\\\\
Consistency (\{min\_range\}-\{max\_range\}) - the factual alignment between the summary and the summarized source. A factually consistent summary contains only statements that are entailed by the source document. Annotators were also asked to penalize summaries that contained hallucinated facts.
\\\\
Evaluation Steps:
\\\\
1. Read the news article carefully and identify the main topic and key points.\\
2. Read the summary and compare it to the news article. Check if the summary covers the main topic and key points of the news article, and if it presents them in a clear and logical order.\\
3. Assign a score for consistency based on the Evaluation Criteria.
\\\\

Example:
\\\\

Source Text:
\\\\

\{\{Document]\}\}
\\\\

Summary:
\\\\

\{\{Summary\}\}
\\\\

Evaluation Form (scores ONLY):
\\\\

- Consistency:
\\\\

What is the consistency of the summary above? Provide only rating and no other text.
\end{promptbox}

\section{Hyper-parameters}
\label{sec:appendix_hyperparameter}
We conduct grid search over two hyperparameters for contrastive decoding: 1) temperature $t$ and 2) scaling constant $\lambda$ from the following ranges:
\begin{itemize}
    \item  $\lambda = [0.01, 0.1, 0.5, 1.0]$
    \item  $t = [0.5, 1.0, 2.0, 3.0, 4.0, 5.0]$
\end{itemize}

The following table shows the hyperparameters setup for each setting:

\begin{table}[h]
\small
\centering
\begin{tabular}{lllll}
\hline
\textbf{Main} & \textbf{Asst} & \textbf{Range} &  \textbf{$\lambda$} & \textbf{$t$} \\
\hline
\multirow{8}{*}{Llama 3.1 8B} & \multirow{4}{*}{Llama 3.2 3B} & 0-4 & 0.01 & 1.0 \\
         &          & 1-5 & 1.0 & 0.5 \\
         &          & 2-6 & 1.0 & 0.5 \\
         &          & 3-7 & 0.01 & 5.0 \\
\cline{2-5}
         & \multirow{4}{*}{Llama 3.2 1B} & 0-4 & 0.01 & 0.5 \\
         &          & 1-5 & 0.1 & 5.0 \\
         &          & 2-6 & 0.1 & 2.0 \\
         &          & 3-7 & 0.1 & 2.0 \\
\hline
\multirow{4}{*}{Qwen 2.5 7B}  & \multirow{4}{*}{Qwen 2.5 3B}  & 0-4 & 0.1 & 4.0 \\
         &          & 1-5 & 0.01 & 5.0 \\
         &          & 2-6 & 0.1 & 4.0 \\
         &          & 3-7 & 0.01 & 0.5 \\
\hline
\multirow{4}{*}{Qwen 2.5 14B} & \multirow{4}{*}{Qwen 2.5 3B}  & 0-4 & 0.1 & 2.0 \\
         &          & 1-5 & 0.01 & 4.0 \\
         &          & 2-6 & 0.1 & 1.0 \\
         &          & 3-7 & 0.1 & 2.0 \\
\hline
\end{tabular}
\caption{Hyperparameter settings for contrastive decoding for each main and assistant model pair from each model family for evaluating summary coherence.}
\label{tab:hyperparameters_coherence}
\end{table}

\begin{table}[h]
\small
\centering
\begin{tabular}{lllll}
\hline
\textbf{Main} & \textbf{Asst} & \bf Range &  \textbf{$\lambda$} & \textbf{$t$} \\
\hline
\multirow{8}{*}{Llama 3.1 8B} & \multirow{4}{*}{Llama 3.2 3B}  & 0-4 & 0.01 & 0.5 \\
         &           & 1-5 & 0.01 & 0.5 \\
         &           & 2-6 & 0.01 & 0.5 \\
         &           & 3-7 & 0.5 & 0.5 \\
         & \multirow{4}{*}{Llama 3.2 1B}  & 0-4 & 0.01 & 0.5 \\
         &           & 1-5 & 0.1 & 5.0 \\
         &           & 2-6 & 0.1 & 5.0 \\
         &           & 3-7 & 0.01  & 0.5 \\
\hline
\multirow{4}{*}{Qwen 2.5 7B}  & \multirow{4}{*}{Qwen 2.5 3B}   & 0-4 & 0.1  & 5.0 \\
         &           & 1-5 & 0.5  & 1.0 \\
         &           & 2-6 & 1.0  & 1.0 \\
         &           & 3-7 & 0.1  & 4.0 \\
\hline
\multirow{4}{*}{Qwen 2.5 14B}& \multirow{4}{*}{Qwen 2.5 3B}   & 0-4 & 0.01 & 3.0 \\
         &           & 1-5 & 0.1  & 0.5 \\
         &           & 2-6 & 0.01 & 3.0 \\
         &           & 3-7 & 0.01 & 3.0 \\
\hline
\end{tabular}
\caption{Hyperparameter settings for contrastive decoding for each Llama-3 main and assistant model pair for relevance.}
\label{tab:hyperparameters_relevance}
\end{table}

\begin{table}[h]
\small
\centering
\begin{tabular}{lllll}
\hline
\textbf{Main} & \textbf{Asst} & \bf Range &  \textbf{$\lambda$} & \textbf{$t$} \\
\hline
\multirow{8}{*}{Llama 3.1 8B} & \multirow{4}{*}{Llama 3.2 3B}  &                      0-4 & 0.01 & 5.0 \\
         &           & 1-5 & 0.1 & 2.0 \\
         &           & 2-6 & 0.1 & 2.0 \\
         &           & 3-7 & 0.1 & 3.0 \\
\cline{2-5}
         & \multirow{4}{*}{Llama 3.2 1B}  & 0-4 & 0.1 & 1.0\\
         &           & 1-5 & 0.1 & 0.5 \\
         &           & 2-6 & 0.1 & 2.0 \\
         &           & 3-7 & 0.1 & 1.0 \\
\hline
\multirow{4}{*}{Qwen 2.5 7B}  & \multirow{4}{*}{Qwen 2.5 3B}   & 0-4 & 0.1 & 3.0 \\
         &           & 1-5 & 0.01 & 5.0 \\
         &           & 2-6 & 0.01 & 2.0 \\
         &           & 3-7 & 0.01 & 5.0 \\
\hline
\multirow{4}{*}{Qwen 2.5 14B}& \multirow{4}{*}{Qwen 2.5 3B}                       & 0-4 & 0.1 & 1.0 \\
         &           & 1-5 & 0.1 & 5.0 \\
         &           & 2-6 & 0.1 & 3.0 \\
         &           & 3-7 & 0.1 & 3.0 \\
\hline
\end{tabular}
\caption{Hyperparameter settings for contrastive decoding for each Llama-3 main and assistant model pair for consistency.}
\label{tab:hyperparameters_consistency}
\end{table}

\section{Score Distribution Across All Ranges}
\label{appendix:score_distributions}

\begin{figure}[t]
    \centering
    \begin{subfigure}{0.49\textwidth}
        \centering
        \includegraphics[width=\textwidth]{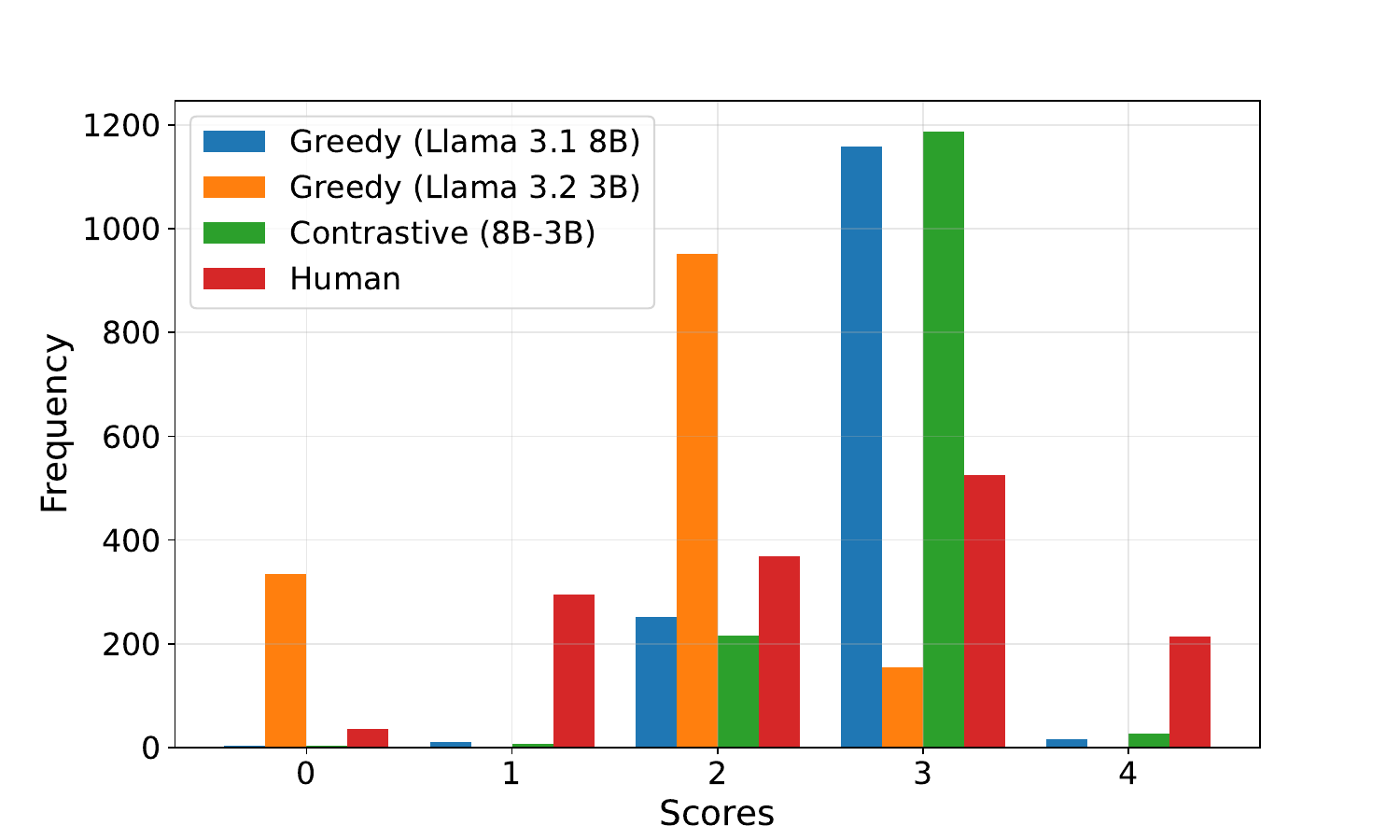}
        \caption{Range 0-4}
        \label{fig:llama_range04}
    \end{subfigure}
    \hfill
    \begin{subfigure}{0.49\textwidth}
        \centering
        \includegraphics[width=\textwidth]{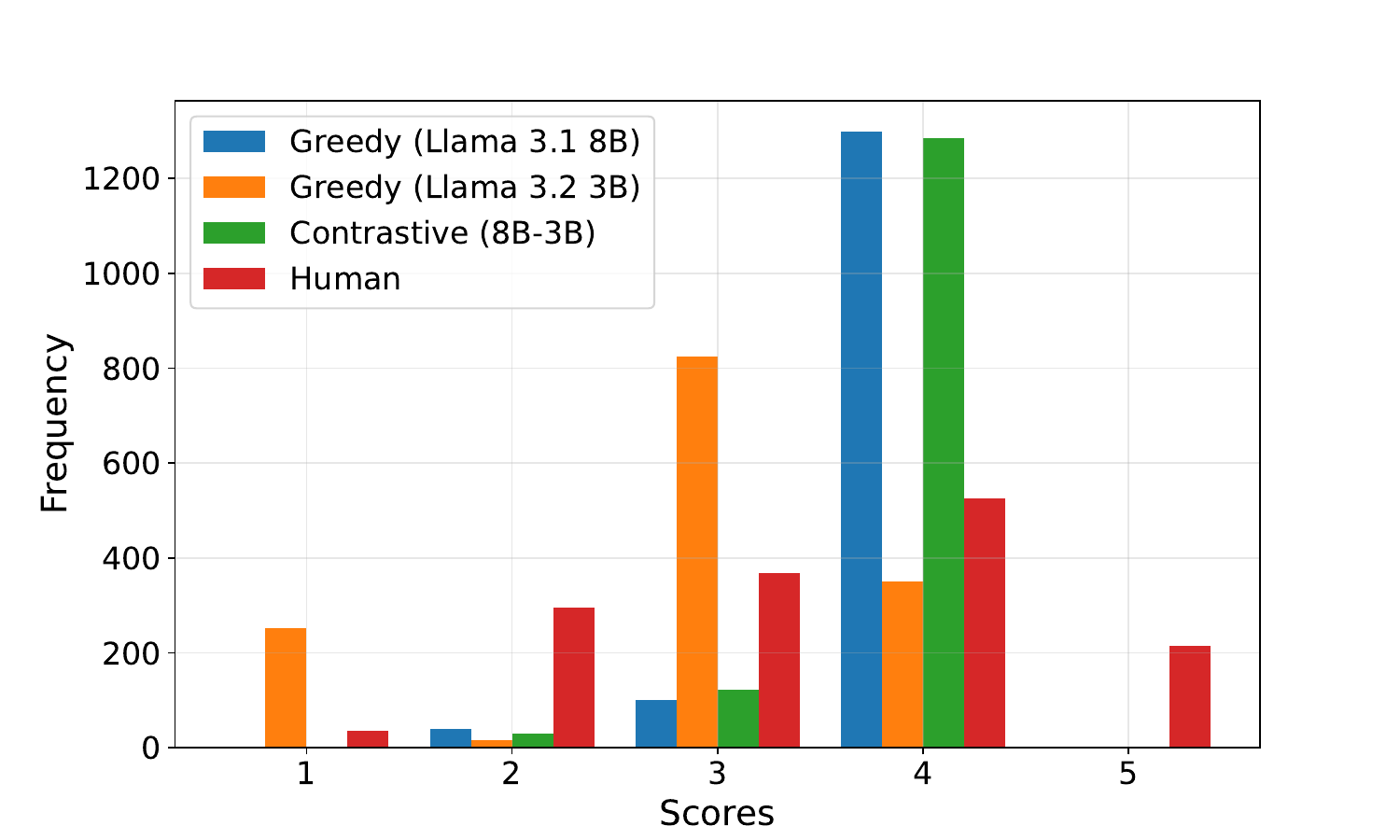}
        \caption{Range 1-5}
        \label{fig:llama_range15}
    \end{subfigure}

    \vspace{0.5cm}

    \begin{subfigure}{0.49\textwidth}
        \centering
        \includegraphics[width=\textwidth]{latex/images/score_distribution_comparison_llama31_8b_coh_detailed_range26.pdf}
        \caption{Range 2-6}
        \label{fig:llama_range26}
    \end{subfigure}
    \hfill
    \begin{subfigure}{0.49\textwidth}
        \centering
        \includegraphics[width=\textwidth]{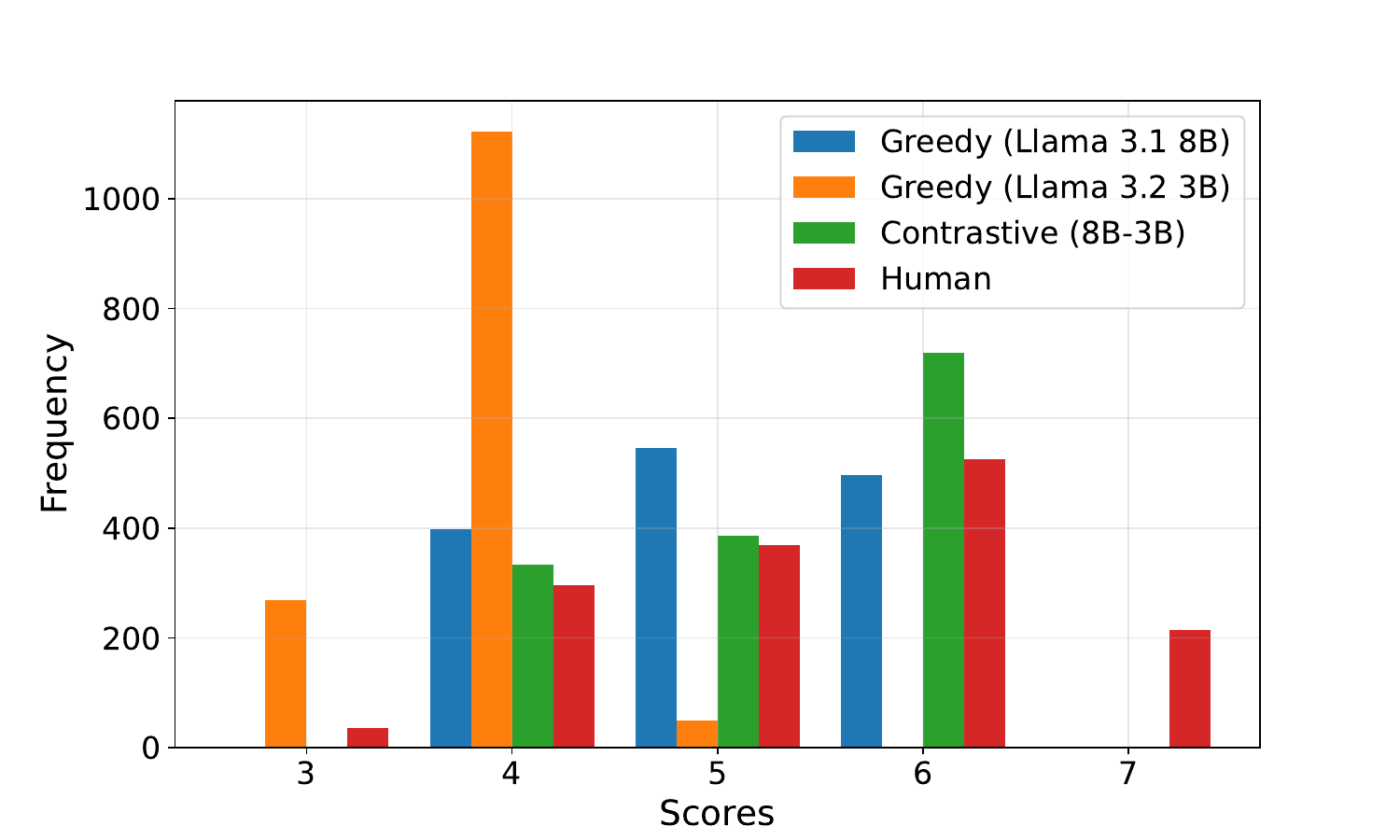}
        \caption{Range 3-7}
        \label{fig:llama_range37}
    \end{subfigure}
    \caption{Predicted score distribution comparison for Llama-3 family across all score ranges on coherence evaluation.}
    \label{fig:llama_all_ranges}
\end{figure}

\begin{figure}[t]
    \centering
    \begin{subfigure}{0.49\textwidth}
        \centering
        \includegraphics[width=\textwidth]{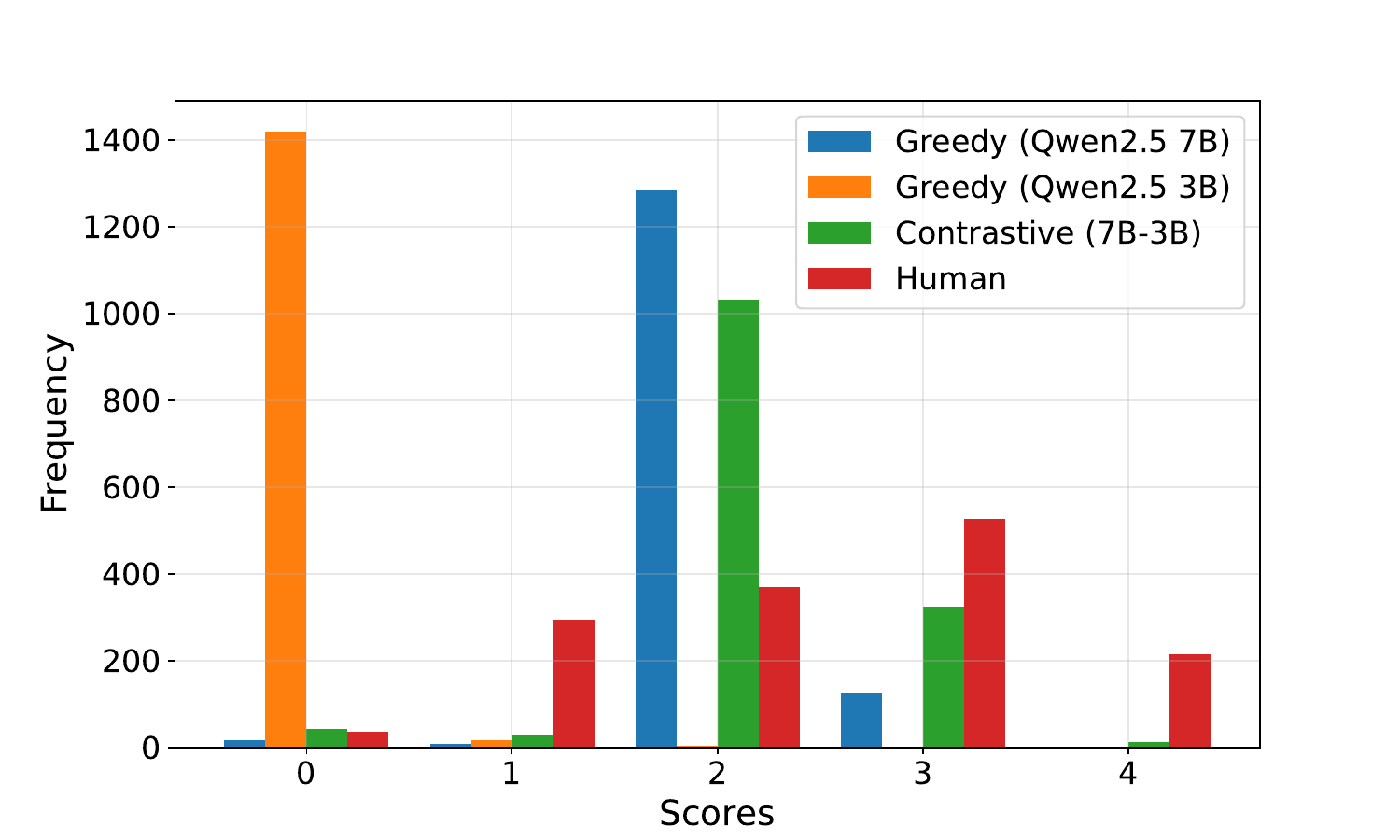}
        \caption{Range 0-4}
        \label{fig:qwen_range04}
    \end{subfigure}
    \hfill
    \begin{subfigure}{0.49\textwidth}
        \centering
        \includegraphics[width=\textwidth]{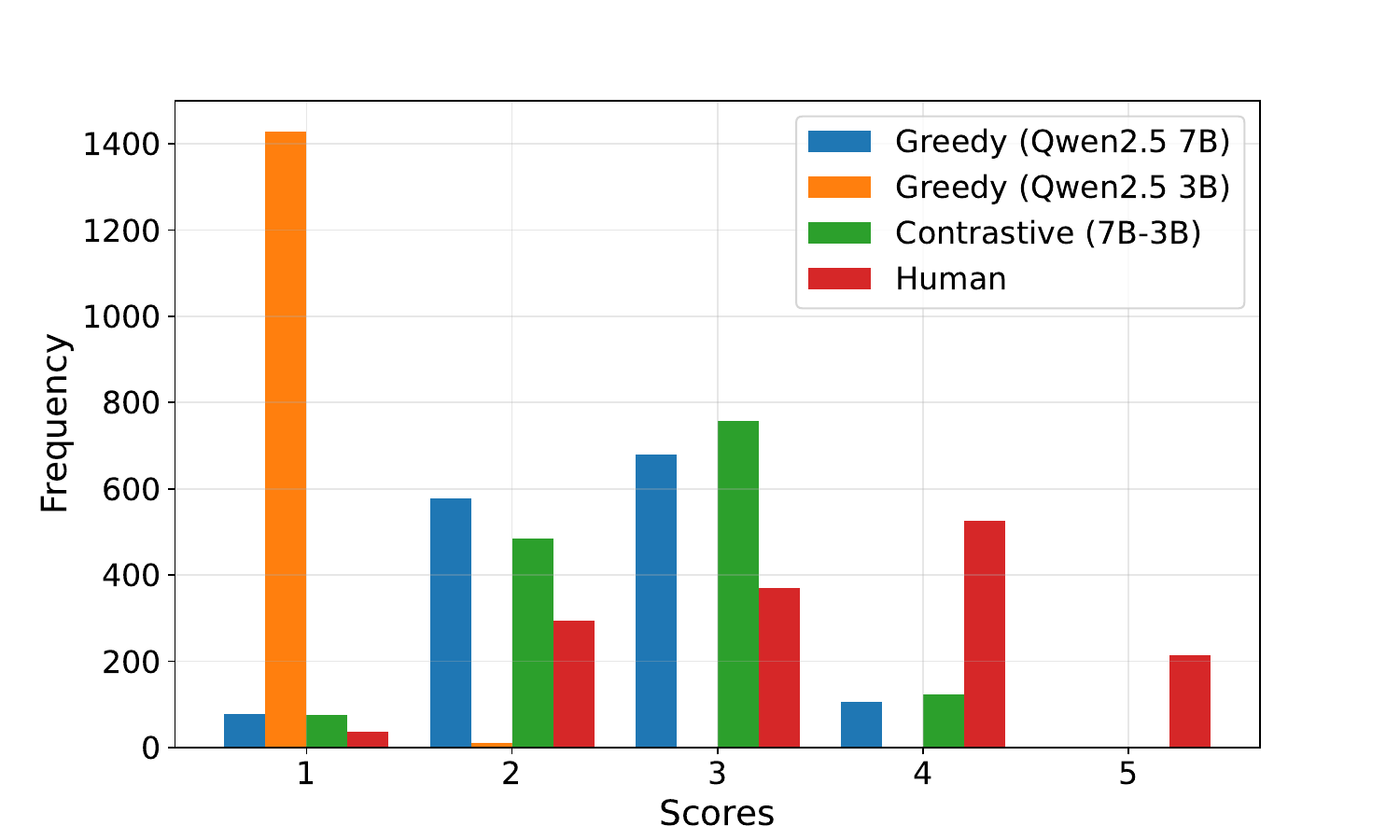}
        \caption{Range 1-5}
        \label{fig:qwen_range15}
    \end{subfigure}

    \vspace{0.5cm}

    \begin{subfigure}{0.49\textwidth}
        \centering
        \includegraphics[width=\textwidth]{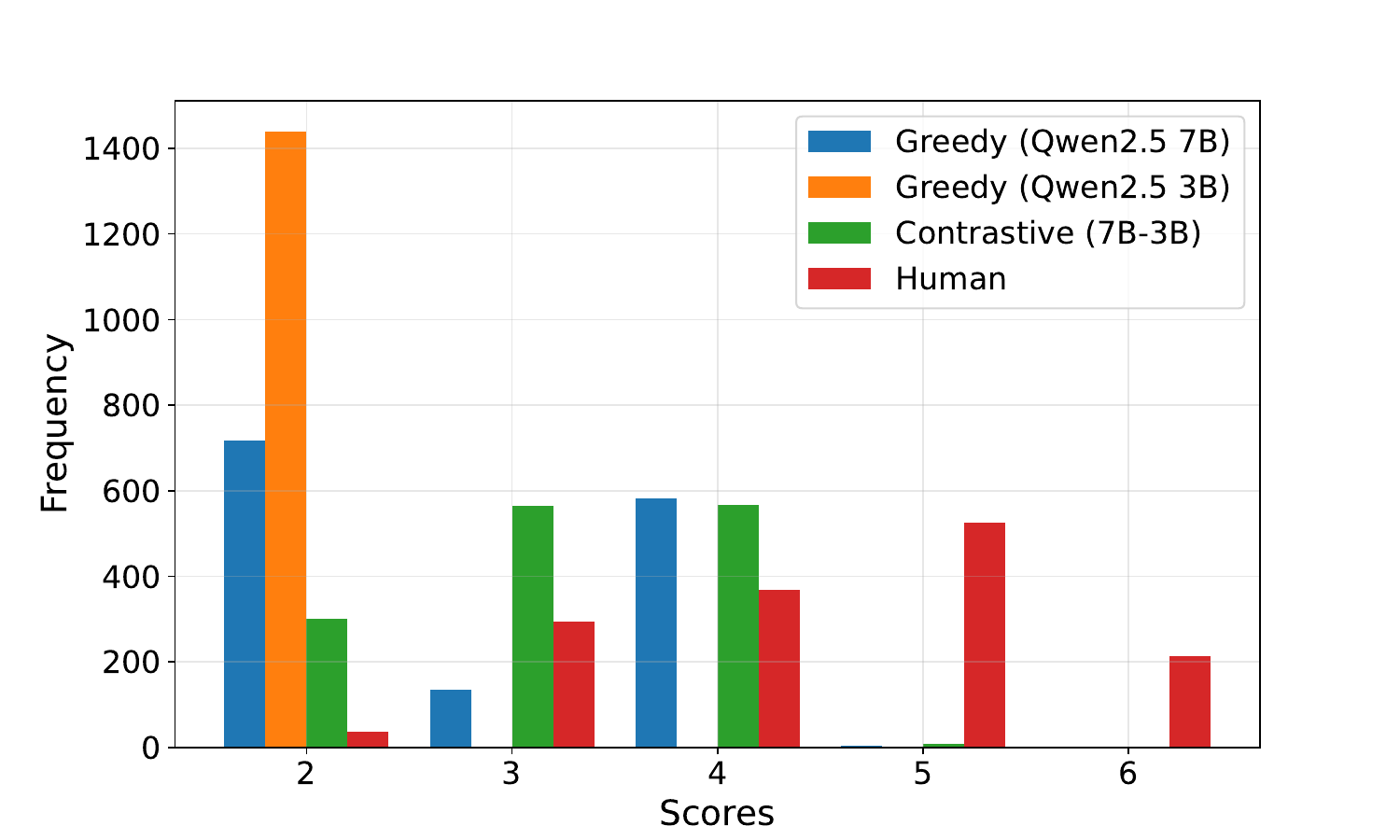}
        \caption{Range 2-6}
        \label{fig:qwen_range26}
    \end{subfigure}
    \hfill
    \begin{subfigure}{0.49\textwidth}
        \centering
        \includegraphics[width=\textwidth]{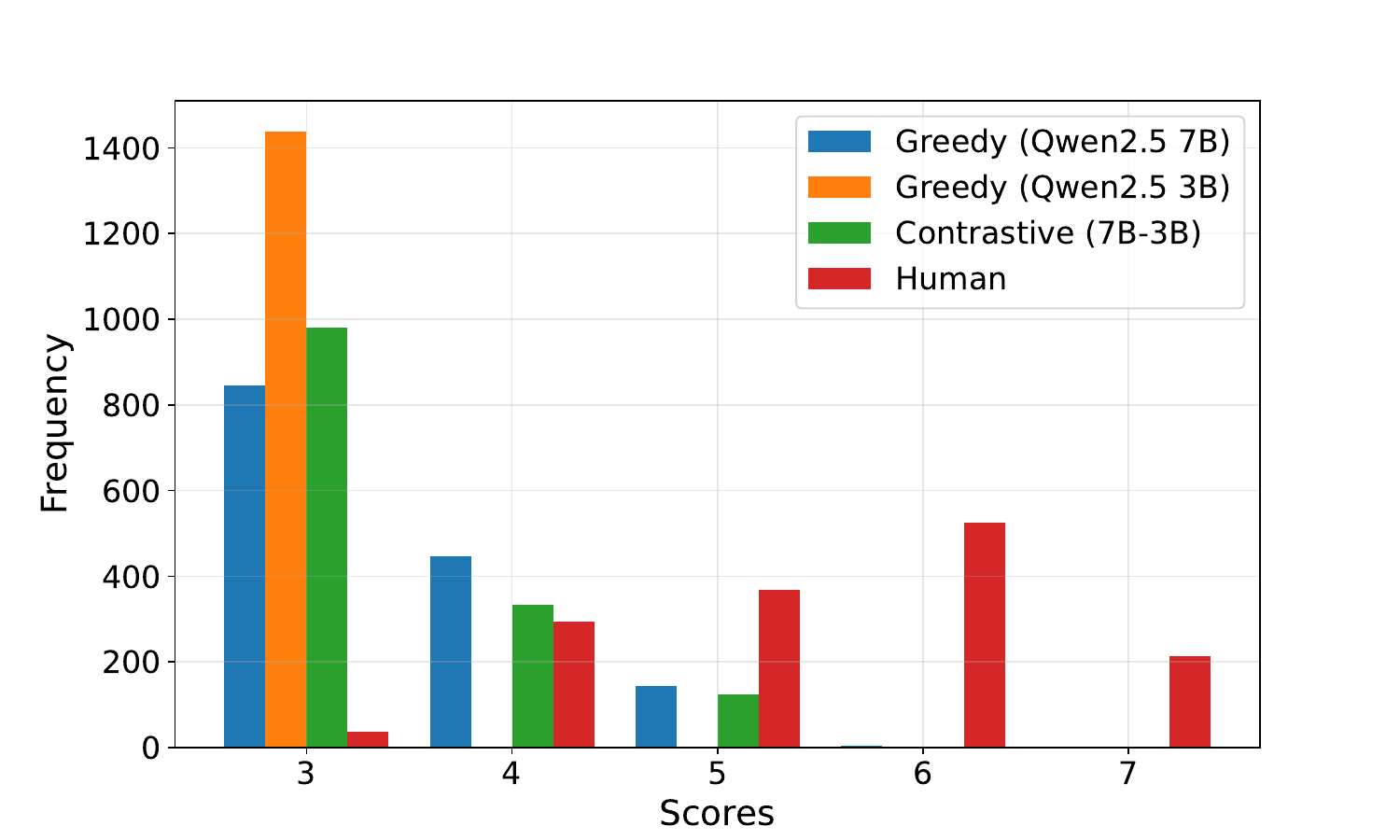}
        \caption{Range 3-7}
        \label{fig:qwen_range37}
    \end{subfigure}
    \caption{Predicted score distribution comparison for Qwen2.5 family across all score ranges on coherence evaluation.}
    \label{fig:qwen_all_ranges}
\end{figure}

\begin{figure}[t]
    \centering
    \begin{subfigure}{0.49\textwidth}
        \centering
        \includegraphics[width=\textwidth]{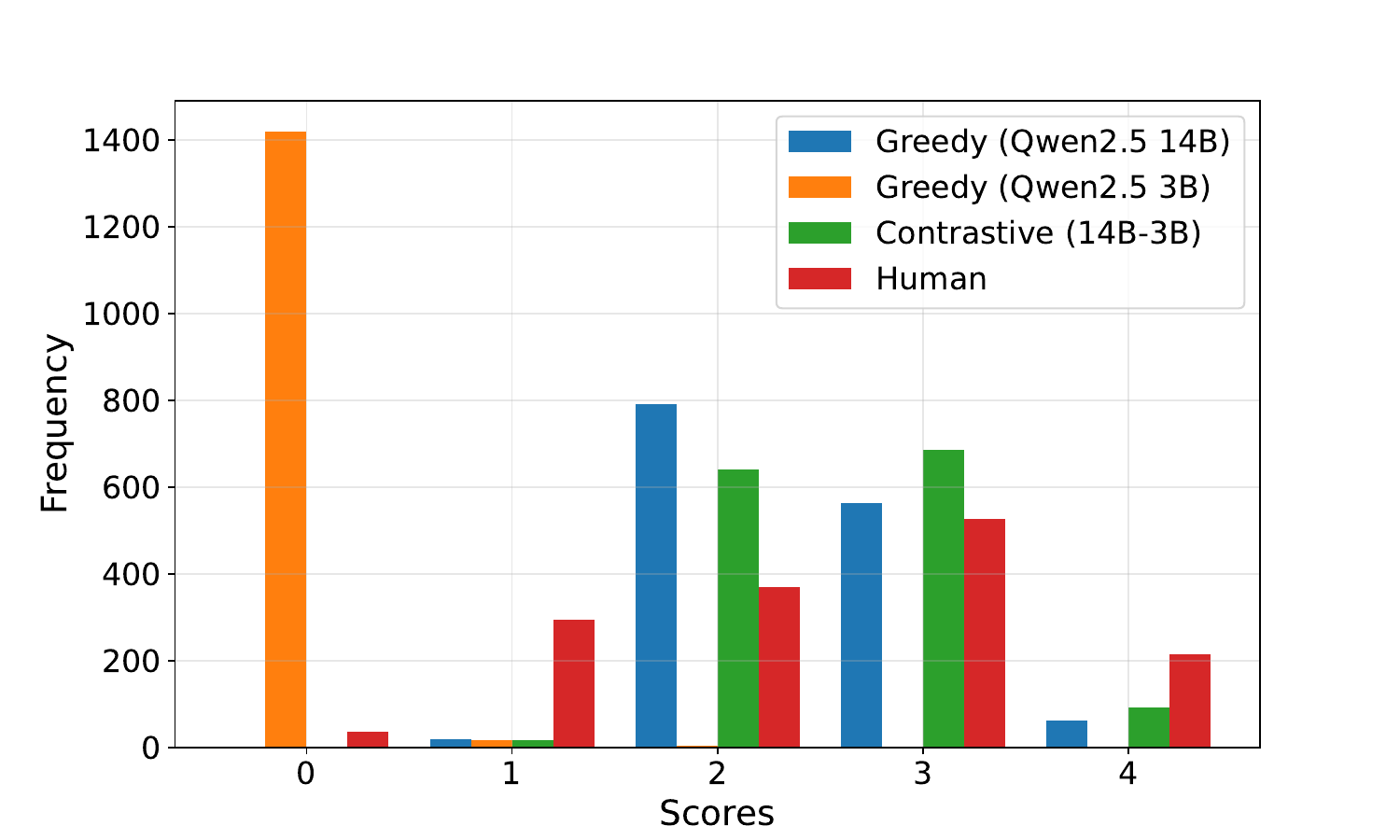}
        \caption{Range 0-4}
        \label{fig:qwen14b_range04}
    \end{subfigure}
    \hfill
    \begin{subfigure}{0.49\textwidth}
        \centering
        \includegraphics[width=\textwidth]{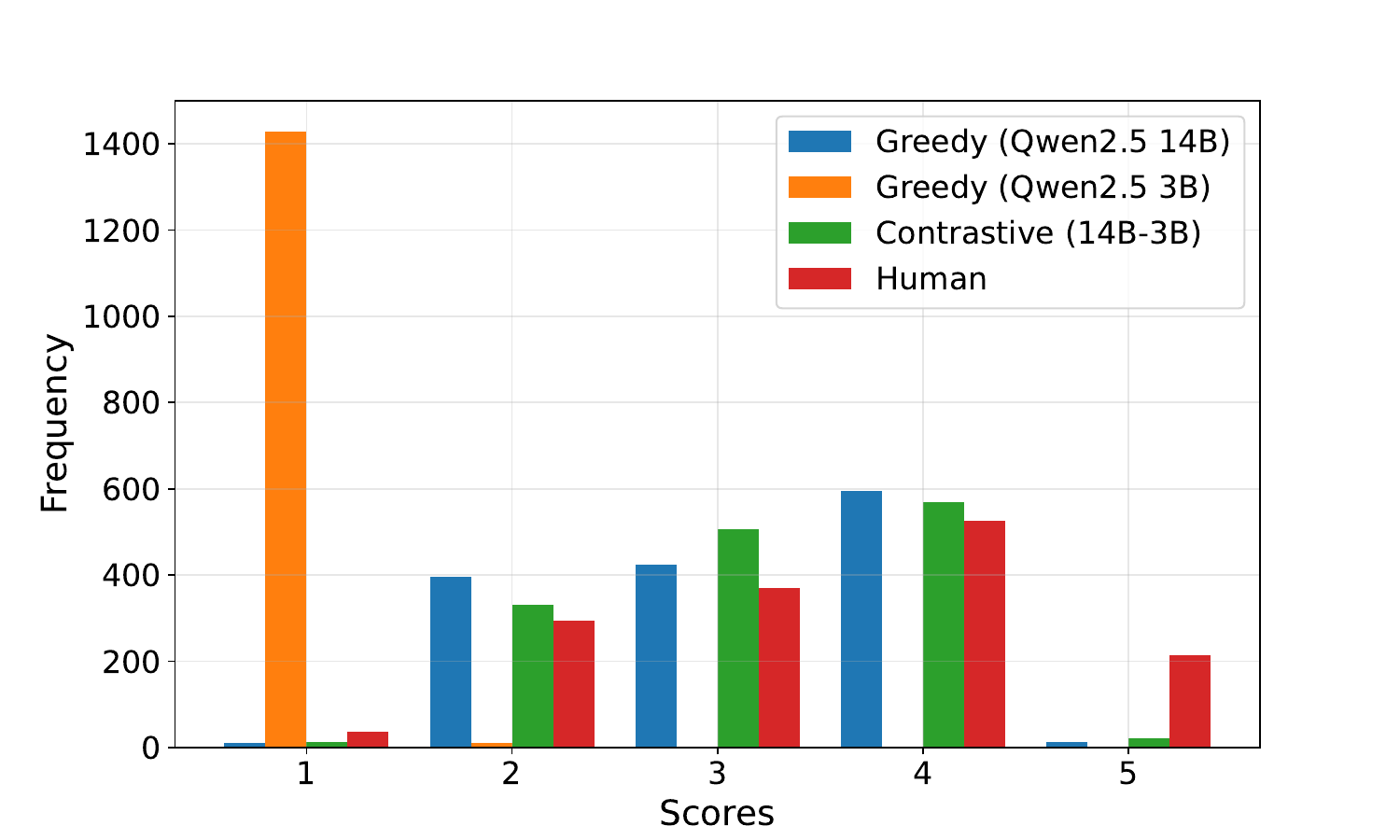}
        \caption{Range 1-5}
        \label{fig:qwen14b_range15}
    \end{subfigure}

    \vspace{0.5cm}

    \begin{subfigure}{0.49\textwidth}
        \centering
        \includegraphics[width=\textwidth]{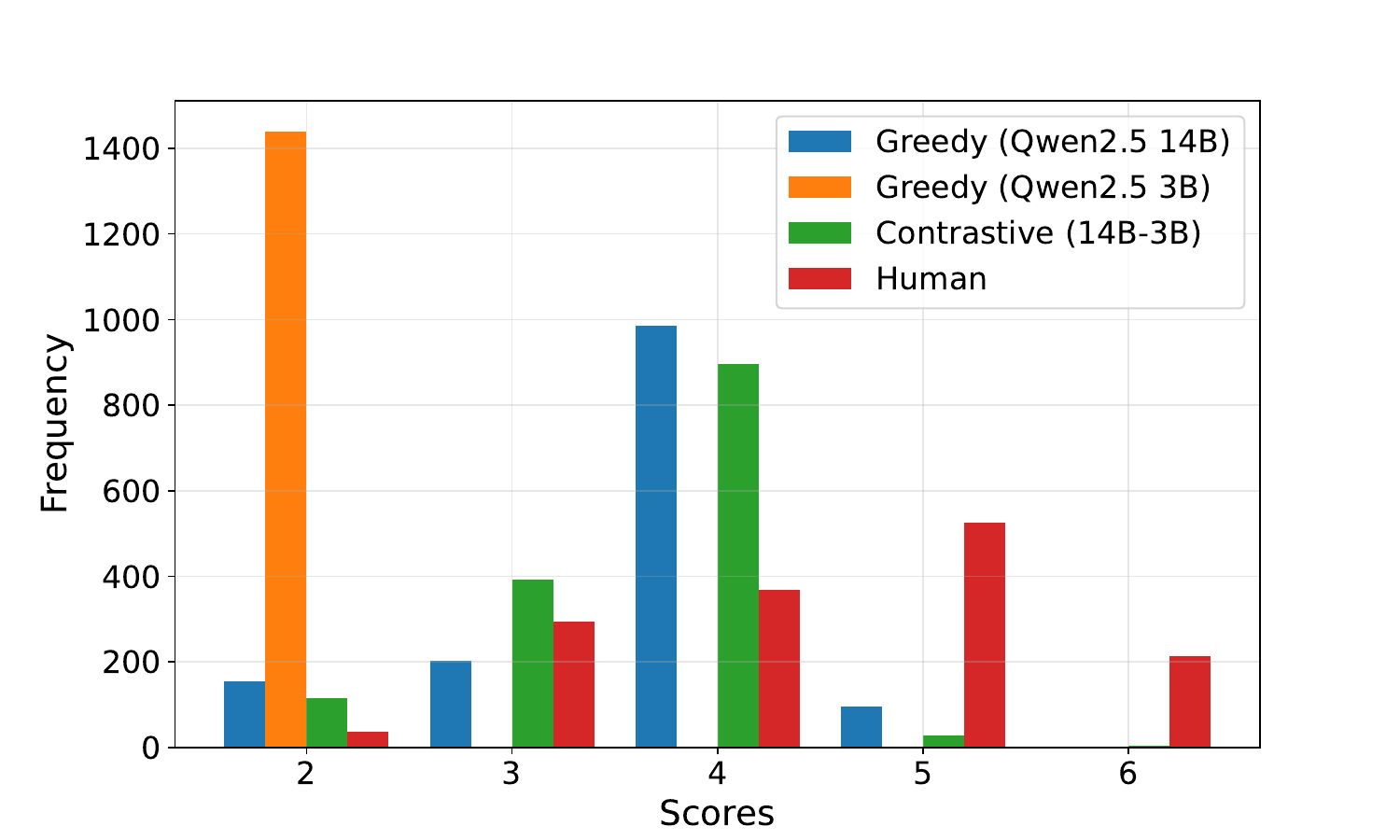}
        \caption{Range 2-6}
        \label{fig:qwen14b_range26}
    \end{subfigure}
    \hfill
    \begin{subfigure}{0.49\textwidth}
        \centering
        \includegraphics[width=\textwidth]{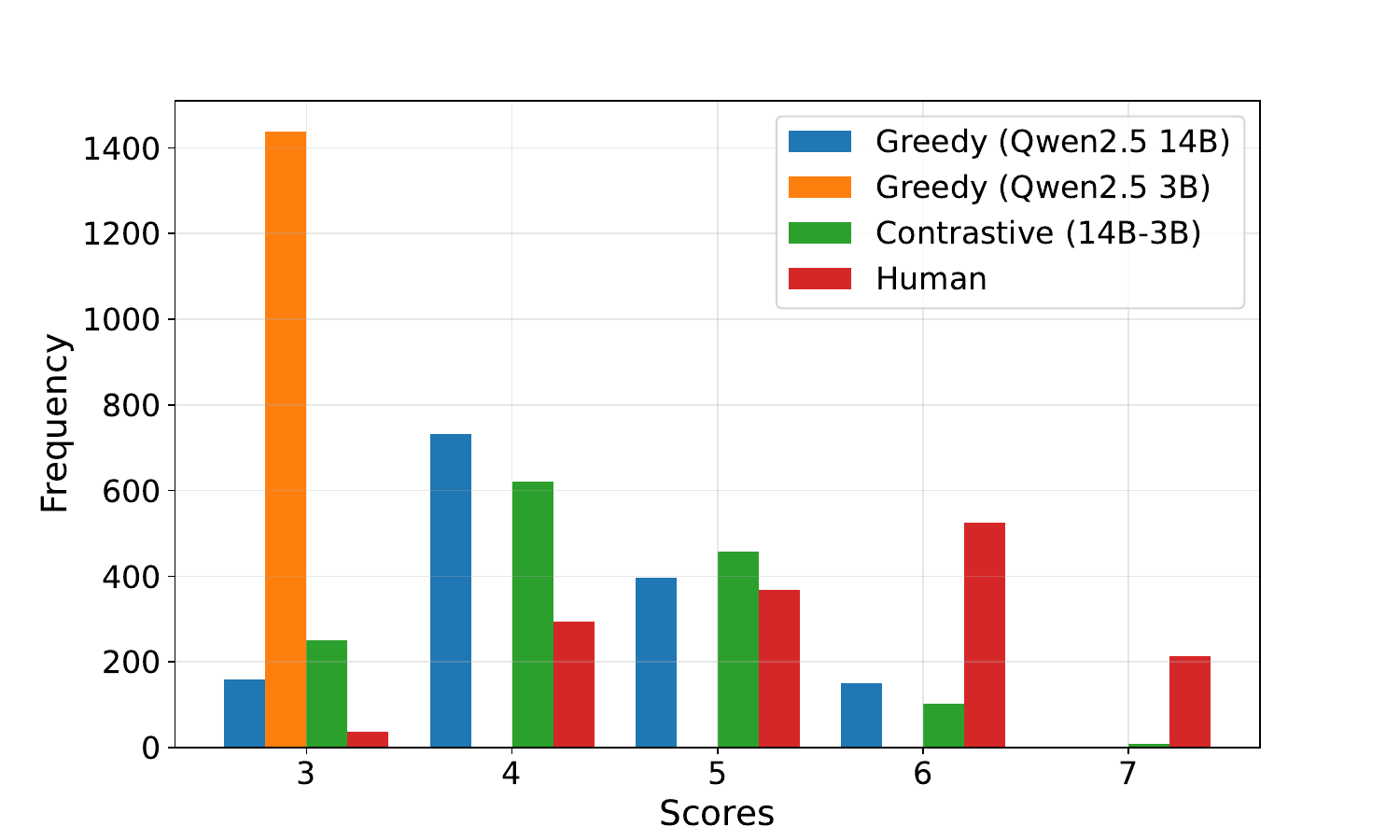}
        \caption{Range 3-7}
        \label{fig:qwen14b_range37}
    \end{subfigure}
    \caption{Score distribution comparison for Qwen2.5 14B family across all score ranges on coherence evaluation.}
    \label{fig:qwen14b_all_ranges}
\end{figure}

Figures~\ref{fig:llama_all_ranges}, \ref{fig:qwen_all_ranges}, and~\ref{fig:qwen14b_all_ranges} show the score distribution comparison plots across all four score ranges (0-4, 1-5, 2-6, 3-7) for Llama-3, Qwen2.5 7B, and Qwen2.5 14B family models respectively. These plots demonstrate how score range bias manifests differently across ranges and model sizes, and how contrastive decoding consistently mitigates this bias.

\subsection{Invalid Output Analysis}
\label{appendix:invalid_outputs}

Tables~\ref{tab:invalid_outputs_llama} and~\ref{tab:invalid_outputs_qwen} show the analysis of invalid model outputs across different score ranges for Llama-3 and Qwen2.5 models. Invalid outputs are classified into three categories: (1) \textit{No Pred}: outputs where no numeric pattern could be extracted, and (2) \textit{Below Range}: outputs with valid numbers below the minimum score for the range.

For Llama models, the 8B model achieves near-perfect parsing with 0\% invalid outputs across all ranges. The 3B model shows moderate failure rates ranging from 17.4\% to 24.0\%, primarily due to malformed outputs. Notably, contrastive decoding maintains excellent performance with minimal failures (0.3\% in the 2-6 range).

For Qwen models, the analysis reveals that the 3B model exhibits catastrophic failure with approximately 98\% invalid outputs across all score ranges, primarily due to malformed outputs that cannot be parsed. In contrast, the 7B and 14B models show significantly better performance, with the 7B model achieving near-perfect parsing (0\% invalid) and the 14B model showing minimal failures, primarily in the 3-7 range where 4.6\% of outputs fall below the minimum score.

\begin{table}[h]
\small
\centering
\begin{tabular}{llrl}
\toprule
Model & Range & Invalid & Failure Types \\
\midrule
\multirow{4}{*}{Llama 3.1 8B}
  & 0-4 & 0.0\% & - \\
  & 1-5 & 0.0\% & - \\
  & 2-6 & 0.0\% & - \\
  & 3-7 & 0.0\% & - \\
\midrule
\multirow{4}{*}{Llama 3.2 3B}
  & 0-4 & 23.2\% & No Pred: 334 \\
  & 1-5 & 17.4\% & No Pred: 251 \\
  & 2-6 & 24.0\% & No Pred: 345 \\
  & 3-7 & 18.6\% & No Pred: 268 \\
\midrule
\multirow{4}{*}{\shortstack[l]{Llama 3.1 8B\\Contrastive}}
  & 0-4 & 0.0\% & - \\
  & 1-5 & 0.0\% & - \\
  & 2-6 & 0.3\% & No Pred: 5 \\
  & 3-7 & 0.0\% & - \\
\bottomrule
\end{tabular}
\caption{Invalid output statistics for Llama-3 models across all score ranges on coherence evaluation. The 8B model shows perfect parsing, while the 3B model exhibits moderate failure rates of 17-24\%.}
\label{tab:invalid_outputs_llama}
\end{table}

\begin{table}[h]
\small
\centering
\begin{tabular}{llrl}
\toprule
Model & Range & Invalid & Failure Types \\
\midrule
\multirow{4}{*}{Qwen2.5 3B}
  & 0-4 & 98.2\% & No Pred: 1414 \\
  & 1-5 & 98.7\% & No Pred: 1421 \\
  & 2-6 & 97.6\% & No Pred: 1406 \\
  & 3-7 & 97.6\% & No Pred: 1376, B: 30 \\
\midrule
\multirow{4}{*}{Qwen2.5 7B}
  & 0-4 & 0.0\% & - \\
  & 1-5 & 0.0\% & - \\
  & 2-6 & 0.0\% & - \\
  & 3-7 & 0.0\% & - \\
\midrule
\multirow{4}{*}{Qwen2.5 14B}
  & 0-4 & 0.0\% & - \\
  & 1-5 & 0.0\% & - \\
  & 2-6 & 0.1\% & B: 1 \\
  & 3-7 & 4.6\% & B: 66 \\
\midrule
\multirow{4}{*}{\shortstack[l]{Qwen2.5 7B\\Contrastive}}
  & 0-4 & 0.0\% & - \\
  & 1-5 & 0.0\% & - \\
  & 2-6 & 0.0\% & - \\
  & 3-7 & 0.0\% & - \\
\midrule
\multirow{4}{*}{\shortstack[l]{Qwen2.5 14B\\Contrastive}}
  & 0-4 & 0.0\% & - \\
  & 1-5 & 0.0\% & - \\
  & 2-6 & 0.1\% & B: 1 \\
  & 3-7 & 5.6\% & B: 81 \\
\bottomrule
\end{tabular}
\caption{Invalid output statistics for Qwen2.5 models across all score ranges on coherence evaluation. The 3B model shows near-complete failure in generating valid scores, while larger models perform significantly better. B: Below}
\label{tab:invalid_outputs_qwen}
\end{table}

\section{Relevance and Consistency Results}
\label{appendix:relevance_consistency}
In Tables~\ref{tab:contrastive_judge_corr_llama_relevance} and~\ref{tab:contrastive_judge_corr_qwen_relevance}, we present the correlation results for summary relevance evaluation across different score ranges for the Llama-3 and Qwen2.5 model families, respectively.
In Tables~\ref{tab:contrastive_judge_corr_llama_consistency} and~\ref{tab:contrastive_judge_corr_qwen_consistency}, we present the correlation results for summary consistency evaluation across different score ranges.
\begin{table}[t]
\small
  \centering
  \label{tab:results_llama_relevance}
  \setlength{\tabcolsep}{4pt}
  \begin{tabular}{lllllllll}
    \toprule
    Model             & Range &  Pear. & Spear. & Kend.   \\
    \midrule
    Llama 3.2-1B   & 0 to 4 & $.028_{.041}$  &  $.028_{.048}$ &  $.024_{.037}$ \\
    Llama 3.2-3B   & 0 to 4 & $.072_{.027}$ & $.111_{.036}$ & $.094_{.025}$ \\
    Llama 3.1-8B   & 0 to 4 & \underline{$.429_{.050}$} &  \underline{$.374_{.027}$} & \underline{$.323_{.016}$} \\
    Contra. (8B-3B) & 0 to 4 & $.407_{.036}$ & $.360_{.038}$ & $.310_{.027}$ \\
    \midrule
    Llama 3.2-1B  & 1 to 5 & $.014_{.054}$ & $.022_{.036}$ & $.019_{.033}$ \\
    Llama 3.2-3B  & 1 to 5 & $.161_{.056}$ & $.179_{.023}$ & $.152_{.023}$ \\
    Llama 3.1-8B  & 1 to 5 & \underline{$.391_{.042}$} & \underline{$.341_{.025}$} & \underline{$.293_{.019}$} \\
    Contra. (8B-3B)& 1 to 5 & $.375_{.030}$ & $.322_{.035}$ & $.277_{.023}$ \\
    \midrule

    Llama 3.2-1B  & 2 to 6 & $.000_{.000}$ & $.000_{.000}$ & $.000_{.000}$ \\
    Llama 3.2-3B  & 2 to 6 & $.076_{.041}$ & $.107_{.037}$ & $.091_{.041}$ \\
    Llama 3.1-8B  & 2 to 6 & \underline{$.378_{.020}$} & \underline{$.360_{.033}$} & \underline{$.312_{.032}$} \\
    Contra. (8B-3B)& 2 to 6 & $.393_{.033}$ & $.381_{.030}$ & $.331_{.025}$ \\
    \midrule
    Llama 3.2-1B  & 3 to 7   & $.036_{.018}$ & $.033_{.018}$ & $.029_{.016}$ \\
    Llama 3.2-3B  & 3 to 7   & $.092_{.044}$ & $.095_{.040}$ & $.082_{.037}$ \\
    Llama 3.1-8B  & 3 to 7   & $.438_{.041}$ & $.421_{.021}$ & $.357_{.019}$ \\
    Contra. (8B-3B)& 3 to 7   & \underline{\it $.471_{.035}$} & \underline{\it $.444_{.023}$} & \underline{\it $.369_{.028}$} \\
    \hline
    \multicolumn{5}{c}{\it Average across all score ranges}  \\
    Llama 3.2-1B & & .026 & .028 & .024  \\
    Llama 3.2-3B & & .100 & .123 & .105 \\
    Llama 3.1-8B & & .409 & .374 & .321 \\
    Contra. (8B-3B)& & \bf{.412} & \bf{.377} & \bf{.322} \\
    \bottomrule
  \end{tabular}
  \caption{\label{tab:contrastive_judge_corr_llama_relevance} Llama-3 family correlation results to human annotations on summary relevance. Max correlation within score range are \underline{underlined}, max across all ranges are {\it italicized}, and max averages are {\bf bolded}.}
\end{table}

\begin{table}[th]
\small
  \centering
  \label{tab:results_qwen}
  \setlength{\tabcolsep}{4pt}
  \begin{tabular}{lllllllll}
    \toprule
    Model             & Range &  Pear. & Spear. & Kend.   \\
    \midrule
    Qwen2.5-3B      & 0 to 4 & $.000_{.000}$ & $.000_{.000}$ & $.000_{.000}$ \\
    Qwen2.5-7B      & 0 to 4 & $.323_{.028}$ & $.296_{.041}$ & $.250_{.045}$ \\
    Qwen2.5-14B     & 0 to 4 & \underline{$.517_{.033}$} & \underline{$.496_{.034}$} &  \underline{$.416_{.023}$} \\
    Contra. (7B-3B)  & 0 to 4 & $.351_{.031}$ & $.289_{.037}$ & $.236_{.053}$ \\
    Contra. (14B-3B) & 0 to 4 & $.516_{.031}$ & $.478_{.038}$ & $.406_{.016}$ \\
    \midrule
    Qwen2.5-3B      & 1 to 5 & $.000_{.000}$ &  $.000_{.000}$ & $.000_{.000}$ \\
    Qwen2.5-7B      & 1 to 5 & $.345_{.037}$ &  $.347_{.043}$ & $.285_{.021}$ \\
    Qwen2.5-14B     & 1 to 5 & \underline{\it $.518_{.036}$} & \underline{\it $.500_{.036}$} & \underline{\it $.427_{.033}$} \\
    Contra. (7B-3B)  & 1 to 5 & $.353_{.045}$ &  $.335_{.036}$ & $.283_{.043}$ \\
    Contra. (14B-3B) & 1 to 5 & $.509_{.024}$ & $.495_{.034}$ &  $.422_{.024}$ \\
    \midrule
    Qwen2.5-3B      & 2 to 6 & $.000_{.000}$ &  $.000_{.000}$ & $.000_{.000}$ \\
    Qwen2.5-7B      & 2 to 6 & $.415_{.040}$ &  $.398_{.024}$ & $.330_{.027}$ \\
    Qwen2.5-14B     & 2 to 6 & $.438_{.034}$ &  $.380_{.038}$ & $.319_{.030}$ \\
    Contra. (7B-3B)  & 2 to 6 & $.171_{.020}$ &  $.134_{.024}$ & $.112_{.027}$ \\
    Contra. (14B-3B) & 2 to 6 & \underline{$.453_{.036}$} &  \underline{$.408_{.027}$} & \underline{$.344_{.024}$} \\
    \midrule
    Qwen2.5-3B      & 3 to 7 & $.000_{.000}$ &  $.000_{.000}$ &  $.000_{.000}$ \\
    Qwen2.5-7B      & 3 to 7 & $.468_{.023}$ &  $.441_{.035}$ &  $.367_{.020}$ \\
    Qwen2.5-14B     & 3 to 7 & $.426_{.034}$ &  $.375_{.037}$ &  $.313_{.023}$ \\
    Contra. (7B-3B)  & 3 to 7 & $.456_{.035}$ &  $.438_{.024}$ &  $.365_{.024}$ \\
    Contra. (14B-3B) & 3 to 7 & \underline{$.510_{.026}$} &  \underline{$.473_{.038}$} &  \underline{$.394_{.016}$} \\
    \hline
    \multicolumn{5}{c}{\it Average across all score ranges}  \\
    Qwen2.5-3B      & & .000 & .000 & .000 \\
    Qwen2.5-7B      & & .388 & .371 & .308 \\
    Qwen2.5-14B     & & .475 & .438 & .369 \\
    Contra. (7B-3B)  & & .333 & .299 & .249 \\
    Contra. (14B-3B) & & \bf{.497} & \bf{.464} & \bf{.392} \\
    \bottomrule
  \end{tabular}
  \caption{\label{tab:contrastive_judge_corr_qwen_relevance} Qwen2.5 family correlation results to human annotations on summary relevance. Max correlation within score range are \underline{underlined} and max averages are {\bf bolded}.}
\end{table}

\begin{table}[t]
\small
  \centering
  \label{tab:results_llama_consistency}
  \setlength{\tabcolsep}{4pt}
  \begin{tabular}{lllllllll}
    \toprule
    Model             & Range &  Pear. & Spear. & Kend.   \\
    \midrule
    Llama 3.2-1B   & 0 to 4 & -$.022_{.025}$ & $.003_{.038}$ & $.003_{.025}$ \\
    Llama 3.2-3B   & 0 to 4 & $.051_{.039}$ & $.073_{.039}$ & $.068_{.034}$ \\
    Llama 3.1-8B   & 0 to 4 & \underline{$.471_{.044}$} & \underline{$.411_{.027}$} & \underline{$.388_{.030}$} \\
    Contra. (8B-3B) & 0 to 4 & $.445_{.039}$ & $.381_{.030}$ & $.360_{.027}$ \\
    \midrule
    Llama 3.2-1B  & 1 to 5 & -$.022_{.007}$ & -$.028_{.007}$ & -$.027_{.009}$ \\
    Llama 3.2-3B  & 1 to 5 & $.178_{.056}$ & $.196_{.038}$ & $.182_{.025}$ \\
    Llama 3.1-8B  & 1 to 5 & \it $.594_{.047}$ & \it $.487_{.053}$ & \it $.465_{.032}$ \\
    Contra. (8B-3B)& 1 to 5 & \underline{$.603_{.056}$} & \underline{$.518_{.048}$} & \underline{$.494_{.035}$} \\
    \midrule

    Llama 3.2-1B  & 2 to 6 & $.000_{.000}$ & $.000_{.000}$ & $.000_{.000}$ \\
    Llama 3.2-3B  & 2 to 6 & $.095_{.054}$ & $.097_{.059}$ & $.092_{.037}$ \\
    Llama 3.1-8B  & 2 to 6 & $.488_{.057}$ & $.431_{.021}$ & $.413_{.045}$ \\
    Contra. (8B-3B)& 2 to 6 & \underline{$.551_{.052}$} & \underline{$.469_{.031}$} & \underline{$.447_{.041}$} \\
    \midrule
    Llama 3.2-1B  & 3 to 7 & -$.011_{.040}$ & -$.004_{.050}$ & -$.004_{.024}$ \\
    Llama 3.2-3B  & 3 to 7 & $.118_{.035}$ & $.136_{.051}$ & $.129_{.038}$ \\
    Llama 3.1-8B  & 3 to 7 & \underline{$.549_{.045}$} & $.484_{.050}$ & $.452_{.050}$ \\
    Contra. (8B-3B)& 3 to 7 & $.540_{.054}$ & \underline{$.496_{.034}$} & \underline{$.465_{.028}$} \\
    \hline
    \multicolumn{5}{c}{\it Average across all score ranges}  \\
    Llama 3.2-1B  & & -.014 & -.007 & -.007 \\
    Llama 3.2-3B  & & .111 & .126 & .118 \\
    Llama 3.1-8B  & & .526 & .453 & .430 \\
    Contra. (8B-3B)& & \bf{.535} & \bf{.466} & \bf{.442} \\
    \bottomrule
  \end{tabular}
  \caption{\label{tab:contrastive_judge_corr_llama_consistency} Llama-3 family correlation results to human annotations on summary consistency. Max correlation within score range are \underline{underlined}, max across all ranges are {\it italicized}, and max averages are {\bf bolded}.}
\end{table}

\begin{table}[th]
\small
  \centering
  \label{tab:results_qwen}
  \setlength{\tabcolsep}{4pt}
  \begin{tabular}{lllllllll}
    \toprule
    Model             & Range &  Pear. & Spear. & Kend.   \\
    \midrule
    Qwen2.5-3B      & 0 to 4 & -$.189_{.085}$ & -$.192_{.062}$ & -$.185_{.051}$ \\
    Qwen2.5-7B      & 0 to 4 & $.396_{.024}$ & $.392_{.021}$ & $.356_{.021}$ \\
    Qwen2.5-14B     & 0 to 4 & \underline{$.422_{.023}$} & \underline{$.443_{.036}$} & \underline{$.410_{.034}$} \\
    Contra. (7B-3B)  & 0 to 4 & $.446_{.025}$ & $.412_{.028}$ & $.369_{.035}$ \\
    Contra. (14B-3B) & 0 to 4 & $.420_{.038}$ & $.437_{.041}$ & $.404_{.035}$ \\
    \midrule
    Qwen2.5-3B      & 1 to 5 & -$.047_{.071}$ & -$.063_{.039}$ & -$.060_{.055}$ \\
    Qwen2.5-7B      & 1 to 5 & \underline{\it $.542_{.044}$} & $.479_{.028}$ & $.444_{.035}$ \\
    Qwen2.5-14B     & 1 to 5 & $.406_{.051}$ & $.417_{.049}$ & $.394_{.037}$ \\
    Contra. (7B-3B)  & 1 to 5 & $.549_{.044}$ & \underline{\it $.474_{.033}$} & \underline{\it $.441_{.030}$} \\
    Contra. (14B-3B) & 1 to 5 & $.477_{.051}$ & $.481_{.045}$ & $.459_{.031}$ \\
    \midrule
    Qwen2.5-3B      & 2 to 6 & -$.114_{.045}$ & -$.095_{.052}$ & -$.091_{.062}$ \\
    Qwen2.5-7B      & 2 to 6 & \underline{$.489_{.035}$} & \underline{$.435_{.032}$} & $.404_{.033}$ \\
    Qwen2.5-14B     & 2 to 6 & $.136_{.034}$ & $.176_{.028}$ & $.162_{.024}$ \\
    Contra. (7B-3B)  & 2 to 6 & $.478_{.021}$ & $.462_{.038}$ & \underline{$.432_{.029}$} \\
    Contra. (14B-3B) & 2 to 6 & $.221_{.029}$ & $.295_{.034}$ & $.269_{.034}$ \\
    \midrule
    Qwen2.5-3B      & 3 to 7 & -$.127_{.074}$ & -$.126_{.033}$ & -$.121_{.049}$ \\
    Qwen2.5-7B      & 3 to 7 & $.391_{.026}$ & \underline{$.414_{.033}$} & \underline{$.367_{.037}$} \\
    Qwen2.5-14B     & 3 to 7 & $.224_{.046}$ & $.204_{.027}$ & $.188_{.030}$ \\
    Contra. (7B-3B)  & 3 to 7 & \underline{$.382_{.028}$} & $.389_{.036}$ & $.345_{.036}$ \\
    Contra. (14B-3B) & 3 to 7 & $.309_{.033}$ & $.269_{.043}$ & $.246_{.034}$ \\
    \hline
    \multicolumn{5}{c}{\it Average across all score ranges}  \\
    Qwen2.5-3B      & & -.119 & -.119 & -.114 \\
    Qwen2.5-7B      & & .455 & .430 & .393 \\
    Qwen2.5-14B     & & .297 & .310 & .289 \\
    Contra. (7B-3B)  & & \bf{.464} & \bf{.434} & \bf{.397} \\
    Contra. (14B-3B) & & .357 & .370 & .345 \\
    \bottomrule
  \end{tabular}
  \caption{\label{tab:contrastive_judge_corr_qwen_consistency} Qwen2.5 family correlation results to human annotations on summary consistency. Max correlation within score range are \underline{underlined}, max across all ranges are {\it italicized}, and max averages are {\bf bolded}.}
\end{table}

\section{Model Size and Budget}
For all the experiments in this paper, NVIDIA's A100 GPU was used. The base models used in this paper are licensed under Meta Llama 3 License\footnote{\url{https://www.llama.com/llama3/license/}} for Llama 3 family models and Apache-2.0 license for Qwen 2.5 family models. We followed their intended use case.

\section{Information About Use of AI Assistants}
We have used Claude on this manuscript to enhance the clarity of the paper and fixing grammatical mistakes. 
We also used it to create the codes to run experiments.

\section{Potential Risks}
As discussed in the limitations section, the experiments are only conducted on English, which may bias the takeaways on English.

\section{BigGen-Bench Results}
\label{appendix:biggen_results}

Tables~\ref{tab:biggen_llama} and~\ref{tab:biggen_qwen} present correlation results on the BigGen-Bench dataset across different score ranges for Llama 3.1 and Qwen2.5 model families.

\begin{table}[th]
\small
  \centering
  \setlength{\tabcolsep}{4pt}
  \begin{tabular}{lllll}
    \toprule
    Model             & Range &  Pear. & Spear. & Kend.   \\
    \midrule
    Llama 3.1-8B     & 0 to 4 & $.384_{.035}$ & $.361_{.034}$ & $.301_{.029}$ \\
    Contra. (8B-1B) & 0 to 4 & \underline{\it $.395_{.036}$} & \underline{\it $.367_{.034}$} & \underline{\it $.307_{.029}$} \\
    \midrule
    Llama 3.1-8B     & 1 to 5 & $.377_{.035}$ & $.345_{.035}$ & $.289_{.030}$ \\
    Contra. (8B-1B) & 1 to 5 & \underline{$.392_{.036}$} & \underline{$.352_{.034}$} & \underline{$.300_{.030}$} \\
    \midrule
    Llama 3.1-8B     & 2 to 6 & $.353_{.037}$ & $.336_{.035}$ & $.284_{.030}$ \\
    Contra. (8B-1B) & 2 to 6 & \underline{$.359_{.035}$} & \underline{$.349_{.034}$} & \underline{$.294_{.029}$} \\
    \midrule
    Llama 3.1-8B     & 3 to 7 & \underline{$.373_{.034}$} & \underline{$.354_{.034}$} & \underline{$.300_{.030}$} \\
    Contra. (8B-1B) & 3 to 7 & $.357_{.037}$ & $.338_{.036}$ & $.288_{.031}$ \\
    \hline
    \multicolumn{5}{c}{\it Average across all score ranges}  \\
    Llama 3.1-8B     & & .372 & .349 & .294 \\
    Contra. (8B-1B) & & \bf{.376} & \bf{.352} & \bf{.297} \\
    \bottomrule
  \end{tabular}
  \caption{\label{tab:biggen_llama} Llama 3.1 family correlation results on BigGen-Bench with 95\% bootstrap confidence intervals. Max correlation within score range are \underline{underlined} and max averages are {\bf bolded}.}
\end{table}

\begin{table}[th]
\small
  \centering
  \setlength{\tabcolsep}{4pt}
  \begin{tabular}{lllll}
    \toprule
    Model             & Range &  Pear. & Spear. & Kend.   \\
    \midrule
    Qwen2.5-14B     & 0 to 4 & $.545_{.031}$ & $.517_{.030}$ & $.445_{.027}$ \\
    Contra. (14B-3B) & 0 to 4 & \underline{\it $.571_{.029}$} & \underline{\it $.533_{.030}$} & \underline{\it $.460_{.026}$} \\
    \midrule
    Qwen2.5-14B     & 1 to 5 & $.520_{.033}$ & $.489_{.032}$ & $.427_{.028}$ \\
    Contra. (14B-3B) & 1 to 5 & \underline{$.534_{.033}$} & \underline{$.498_{.032}$} & \underline{$.435_{.028}$} \\
    \midrule
    Qwen2.5-14B     & 2 to 6 & $.533_{.032}$ & $.501_{.032}$ & $.436_{.028}$ \\
    Contra. (14B-3B) & 2 to 6 & \underline{$.556_{.031}$} & \underline{$.530_{.031}$} & \underline{$.459_{.027}$} \\
    \midrule
    Qwen2.5-14B     & 3 to 7 & $.528_{.033}$ & $.494_{.033}$ & $.432_{.029}$ \\
    Contra. (14B-3B) & 3 to 7 & \underline{$.539_{.032}$} & \underline{$.507_{.030}$} & \underline{$.443_{.028}$} \\
    \hline
    \multicolumn{5}{c}{\it Average across all score ranges}  \\
    Qwen2.5-14B     & & .532 & .500 & .435 \\
    Contra. (14B-3B) & & \bf{.550} & \bf{.517} & \bf{.449} \\
    \bottomrule
  \end{tabular}
  \caption{\label{tab:biggen_qwen} Qwen2.5 family correlation results on BigGen-Bench with 95\% bootstrap confidence intervals. Qwen2.5 14B with contrastive decoding (14B-3B) consistently achieves the best performance across all score ranges. Max correlation within score range are \underline{underlined}, and max averages are {\bf bolded}.}
\end{table}

\section{Assistant Model Size Analysis}
\label{appendix:assistant_size}

The paper notes that smaller assistant models sometimes work better than larger ones, but does not explore why. We conducted a follow-up experiment using Qwen2.5-7B as an assistant model (instead of the 3B model) for the Qwen2.5-14B main model on summary coherence evaluation.

Table~\ref{tab:qwen14b_7b_coherence} shows the results for Qwen 14B-7B contrastive decoding across all score ranges. We observe no significant changes in the 2-6 and 3-7 ranges compared to using the 3B assistant. However, there is a notable degradation in the 1-5 range, where the 7B assistant model significantly reduces correlation (Spearman: 0.282 vs 0.470 with 3B assistant, see Table~\ref{tab:contrastive_judge_corr_qwen}). This finding supports the hypothesis that using a larger assistant model is more likely to penalize correct logits from the main model, thereby degrading performance.

\begin{table}[h]
\small
  \centering
  \setlength{\tabcolsep}{4pt}
  \begin{tabular}{llll}
    \toprule
    Range & Pearson & Spearman & Kendall \\
    \midrule
    0-4 & $.419_{.026}$ & $.444_{.040}$ & $.370_{.027}$ \\
    1-5 & $.293_{.027}$ & $.282_{.023}$ & $.229_{.035}$ \\
    2-6 & $.402_{.027}$ & $.412_{.050}$ & $.336_{.031}$ \\
    3-7 & $.418_{.034}$ & $.402_{.026}$ & $.330_{.034}$ \\
    \hline
    \multicolumn{4}{c}{\it Average across all score ranges} \\
    & .383 & .385 & .316 \\
    \bottomrule
  \end{tabular}
  \caption{\label{tab:qwen14b_7b_coherence} Qwen2.5 14B-7B contrastive decoding results on summary coherence with 95\% bootstrap confidence intervals. The 1-5 range shows significantly degraded performance compared to using a 3B assistant (Table~\ref{tab:contrastive_judge_corr_qwen}), supporting the hypothesis that larger assistant models can penalize correct logits.}
\end{table}

\end{document}